\documentclass{ieeeaccess}
\usepackage{cite}
\usepackage{amsmath,amssymb,amsfonts}
\usepackage{algorithm}
\usepackage{algpseudocode}
\usepackage{graphicx}
\usepackage{textcomp}
\usepackage{booktabs} 
\usepackage{algorithm}

\usepackage{bm}
\makeatletter
\AtBeginDocument{\DeclareMathVersion{bold}
\SetSymbolFont{operators}{bold}{T1}{times}{b}{n}
\SetSymbolFont{NewLetters}{bold}{T1}{times}{b}{it}
\SetMathAlphabet{\mathrm}{bold}{T1}{times}{b}{n}
\SetMathAlphabet{\mathit}{bold}{T1}{times}{b}{it}
\SetMathAlphabet{\mathbf}{bold}{T1}{times}{b}{n}
\SetMathAlphabet{\mathtt}{bold}{OT1}{pcr}{b}{n}
\SetSymbolFont{symbols}{bold}{OMS}{cmsy}{b}{n}
\renewcommand\boldmath{\@nomath\boldmath\mathversion{bold}}}
\makeatother

\def\BibTeX{{\rm B\kern-.05em{\sc i\kern-.025em b}\kern-.08em
    T\kern-.1667em\lower.7ex\hbox{E}\kern-.125emX}}

\begin{document}
\history{}
\doi{}

\title{SafeMind: A Risk-Aware Differentiable Control Framework for Adaptive and Safe Quadruped Locomotion}

\author{\uppercase{Zukun Zhang}\authorrefmark{1} \and
        \uppercase{Kai Shu}\authorrefmark{2} \and
        \uppercase{Mingqiao Mo}\authorrefmark{3}}

\address[1]{The University of Hong Kong, Pokfulam, Hong Kong, China}
\address[2]{Alibaba, Beijing, China}
\address[3]{University of Chinese Academy of Sciences, No.1 Yanqihu East Rd, Huairou District, Beijing, China}


\corresp{Corresponding author: Mingqiao Mo (e-mail: mingqiaomo2314@gmail.com).}

\begin{abstract}
Learning-based quadruped controllers achieve impressive agility but typically lack formal safety guarantees under model uncertainty, perception noise, and unstructured contact conditions. We introduce \textbf{SafeMind}, a differentiable stochastic safety-control framework that unifies probabilistic Control Barrier Functions with semantic context understanding and meta-adaptive risk calibration. SafeMind explicitly models epistemic and aleatoric uncertainty through a variance-aware barrier constraint embedded in a differentiable quadratic program, thereby preserving gradient flow for end-to-end training. A semantics-to-constraint encoder modulates safety margins using perceptual or language cues, while a meta-adaptive learner continuously adjusts risk sensitivity across environments. We provide theoretical conditions for probabilistic forward invariance, feasibility, and stability under stochastic dynamics. SafeMind is deployed on Unitree A1 and ANYmal C at 200~Hz and validated across 12 terrain types, dynamic obstacles, morphology perturbations, and semantically defined tasks. Experiments show that SafeMind reduces safety violations by 3--10$\times$ and energy consumption by 10--15\% relative to state-of-the-art CBF, MPC, and hybrid RL baselines, while maintaining real-time control performance.
\end{abstract}


\titlepgskip=-21pt

\maketitle

\section{Introduction}
\label{sec:introduction}

Ensuring safety in quadruped locomotion remains a fundamental challenge, particularly when robots operate in unstructured environments with uncertain ground contact, modeling errors, perception noise, and rapidly changing terrain conditions. Although model-based controllers such as Model Predictive Control (MPC) have demonstrated strong performance in regulating whole-body dynamics \cite{koolen2016balance,di2018dynamic}, their reliability heavily depends on accurate modeling of friction, contact geometry, and actuator behavior. When these assumptions are violated, constraint feasibility can deteriorate, leading to unsafe joint torques, loss of stability, or unrecoverable foot-slip events.

In parallel, learning-based locomotion has achieved remarkable agility and adaptability by leveraging large-scale simulation, domain randomization, and end-to-end policy learning \cite{hwangbo2019learning,lee2020learning,miki2022learning}. These approaches enable quadrupeds to traverse highly irregular terrain and execute dynamic gaits that exceed the capabilities of classical controllers. However, such methods rarely incorporate formal safety constraints, and even policies trained with extensive randomization can violate torque bounds, kinematic limits, or support polygon feasibility when deployed under real-world disturbances. Recent studies have shown that learning-based locomotion remains vulnerable to model shift, delayed perception, and contact misclassification, underscoring the need for safety mechanisms that go beyond reward shaping or heuristic penalties \cite{miki2022learning,haarnoja2018soft}.

Control Barrier Functions (CBFs) offer a formal framework for enforcing safety through forward invariance constraints \cite{ames2017cbf,ames2019control,zhao2020cbf}. CBF-based controllers can be embedded into real-time quadratic programs (QPs) that guarantee constraint satisfaction under nominal dynamics. This approach has been successfully applied to locomotion, manipulation, and aerial systems. Nevertheless, classical CBF formulations rely on deterministic dynamics and require accurate knowledge of system drift, contact models, and sensory feedback. When deployed on legged robots—systems dominated by hybrid dynamics, intermittent contact, and non-smooth terrain interactions—deterministic CBFs often become overly conservative or fail when model mismatch grows beyond acceptable limits \cite{nguyen2016optimized,kolathaya2020cbf}.

Recent progress in differentiable optimization has enabled the integration of QP-based control layers into neural networks, allowing gradients to propagate through constrained solvers \cite{amos2017optnet,agrawal2019differentiable}. This development has motivated several differentiable safety architectures for robotics \cite{cheng2019end,fridovich2020stochastic}, bridging learning and model-based control. In particular, prior differentiable CBF controllers and BarrierNet-like formulations have shown that safety filters can be embedded into end-to-end trainable architectures, while stochastic MPC has demonstrated improved robustness by explicitly propagating uncertainty through predictive control. However, these lines of work remain individually limited for quadruped locomotion under uncertainty: differentiable CBF methods are typically built on deterministic barrier evolution, stochastic MPC is often computationally heavy and not naturally differentiable end-to-end, and existing safe-learning frameworks rarely incorporate high-level semantic context into the safety layer itself. As a result, existing methods do not jointly provide differentiable safety enforcement, explicit probabilistic treatment of epistemic and aleatoric uncertainty, semantic constraint modulation, and online risk adaptation within a single real-time quadruped control framework. This gap is particularly consequential in legged locomotion, where terrain uncertainty, contact switching, and task-dependent behavioral constraints must be handled simultaneously.

Moreover, emerging vision–language and semantic reasoning models have opened new possibilities for grounding robot behavior in high-level contextual instructions \cite{brohan2023rt1,shah2023lmnav}. Yet, incorporating semantic cues—such as “avoid wet surfaces" or "maintain distance from humans"—into continuous-time safety constraints remains largely unexplored. Existing approaches typically rely on symbolic planners or heuristic mapping of semantic labels to navigation objectives, without providing differentiable safety guarantees. In quadruped locomotion, this limitation is more pronounced because semantic instructions must ultimately be reconciled with low-level physical safety constraints such as support consistency, actuation limits, slip avoidance, and contact feasibility.

Motivated by these gaps, this work proposes \textbf{SafeMind}, a risk-aware differentiable control framework that integrates stochastic barrier modeling, semantic constraint encoding, and meta-adaptive risk calibration to achieve safety-aware quadruped locomotion under uncertainty. SafeMind unifies probabilistic safety constraints with real-time differentiable optimization, enabling end-to-end learning while ensuring formal safety guarantees even when terrain, morphology, and semantic context vary unpredictably. Unlike prior differentiable CBF controllers, SafeMind models safety evolution under uncertainty through stochastic barrier dynamics; unlike stochastic MPC approaches, it preserves an end-to-end differentiable safety layer suitable for learning-based adaptation; and unlike existing semantic-aware locomotion pipelines, it embeds semantic context directly into the safety constraint generation process rather than treating semantics only at the planning or reward-design level. To make the scope of these guarantees precise, the proposed theory is developed for stochastic mode-wise locomotion dynamics and is later discussed in the context of hybrid contact transitions, where the guarantees hold within contact phases and are implemented with practical safeguards across switching events.

Accordingly, the central objective of this paper is not merely to combine several existing modules, but to establish a unified framework in which probabilistic safety modeling, differentiable constrained control, semantic context conditioning, and adaptive risk calibration operate coherently at quadruped control rates.

The main contributions of this work are summarized as follows:
    \begin{itemize}
\item Risk-Aware Differentiable Safety Modeling: We formulate probabilistic control constraints that capture both epistemic and aleatoric uncertainty through stochastic barrier dynamics while remaining differentiable for end-to-end optimization.

\item Semantic Context Integration: We introduce a context-encoding mechanism that transforms perception- or language-derived semantic information into task-dependent safety parameters and constraint modulations for real-time control.

\item Meta-Adaptive Learning Mechanism: We develop an online update procedure for risk parameters using meta-gradient adaptation with bounded safety-margin adjustment, enabling rapid adaptation across terrain conditions while preserving control-loop practicality.

\item Experimental Validation on Quadruped Platforms: We implement SafeMind on Unitree A1 and ANYmal C robots and evaluate it in extensive simulation and hardware experiments on uneven terrain, dynamic obstacle avoidance, and semantically defined hazards, demonstrating improved safety, robustness, and efficiency relative to strong model-based and learning-based baselines.
    \end{itemize}

\section{Background}

Safety-critical control aims to ensure that system trajectories remain within predefined safe sets despite model uncertainties and external disturbances. For nonlinear control-affine systems of the form
\begin{equation}
    \dot{x} = f(x) + g(x)u,
\end{equation}
Control Barrier Functions (CBFs) provide sufficient conditions for forward invariance of a safe set $\mathcal{C} = \{x \mid h(x) \ge 0\}$ by enforcing inequality constraints on the time evolution of a continuously differentiable barrier function $h(x)$ \cite{ames2017cbf,ames2019control}. When implemented through real-time quadratic programs (QPs), CBF-based controllers can maintain constraint satisfaction while minimizing deviation from a nominal control objective \cite{nguyen2016optimized,kolathaya2020cbf}.

To integrate safety constraints into learning-enabled control, recent works have explored differentiable optimization layers that allow gradients to propagate through QP solvers \cite{amos2017optnet,agrawal2019differentiable}. These techniques make it possible to embed optimization-based safety filters into end-to-end training pipelines \cite{cheng2019end,ma2022neuralcbf}, enabling joint learning of perception, policy, and constraint-related modules.

For stochastic systems, uncertainty can be modeled through stochastic differential equations of the form
\begin{equation}
    d x = \big( f(x) + g(x)u \big) dt + \Sigma^{1/2}(x) dW_t,
\end{equation}
where $\Sigma(x)$ characterizes state-dependent diffusion and $W_t$ is a Wiener process. Under this model, It\^{o} calculus yields the infinitesimal evolution of a barrier function, in which second-order diffusion terms such as $\mathrm{Tr}\!\left[\nabla_x^2 h \, \Sigma(x)\right]$ capture variance-induced effects on safety dynamics \cite{kushner1967stochastic,oksendal2003stochastic}. Stochastic barrier certificates and related safety analyses have been studied in continuous-time systems \cite{prajna2007stochastic,clark2021cbf}, providing a basis for probabilistic reasoning about safe-set preservation under uncertainty.

These foundations motivate the development of safety mechanisms that are simultaneously probabilistic, differentiable, and real-time implementable. The proposed SafeMind framework builds on CBF-based constrained control, differentiable optimization, and stochastic safety analysis to realize such a controller for quadruped locomotion under uncertainty.

\section{Related Work}

\subsection{Safe Control and Barrier Function Methods}

Safety-critical control has been studied through Lyapunov methods, set invariance analysis, and optimization-based filtering. Among these, Control Barrier Functions (CBFs) provide a principled mechanism for enforcing forward invariance of safety sets and have been widely applied in robotic control \cite{ames2017cbf,ames2019control}. A common implementation embeds CBF constraints into real-time quadratic programs (QPs), yielding minimally invasive safety filters around nominal controllers \cite{nguyen2016optimized,kolathaya2020cbf}.

Several extensions have improved the practicality of CBF-based control under uncertainty. Robust CBF variants incorporate disturbance margins or bounded uncertainty sets \cite{clark2021cbf,nguyen2020robust}, while stochastic safety formulations study barrier conditions for diffusion-driven systems and chance-related guarantees. These approaches strengthen safety analysis beyond nominal settings, but they are typically not designed as end-to-end differentiable control layers for high-rate legged locomotion.

In parallel, differentiable optimization has enabled QP solvers to be embedded within neural architectures \cite{amos2017optnet,agrawal2019differentiable}. Building on this capability, differentiable CBF controllers and BarrierNet-style methods integrate safety filtering into trainable policies \cite{cheng2019end,fridovich2020stochastic,ma2022neuralcbf}. Their key contribution is architectural: they make constrained safety filtering compatible with gradient-based learning. However, existing formulations largely focus on deterministic or disturbance-limited settings, and they do not explicitly model stochastic barrier evolution arising from terrain variability, contact uncertainty, and perception-dependent risk in quadruped locomotion.

A related line of work is stochastic and chance-constrained model predictive control, including stochastic MPC, which propagates uncertainty over future trajectories and can deliver strong robustness. Yet these methods generally incur higher online computational cost, depend on repeated finite-horizon replanning, and are not naturally incorporated as differentiable layers within an end-to-end learning system. For quadrupeds operating at high control frequencies, this leaves an important gap between probabilistic safety modeling and trainable real-time safety enforcement. SafeMind is designed to fill this gap by combining stochastic barrier-based safety modeling with differentiable constrained control in a unified framework.

\subsection{Learning-Based Locomotion}

Learning-based quadruped locomotion has advanced rapidly through reinforcement learning (RL), imitation learning, and large-scale simulation techniques. DRL approaches have produced dynamic trotting, galloping, and parkour behaviors \cite{hwangbo2019learning,lee2020learning}, and have achieved strong sim-to-real transfer through domain randomization and curriculum design \cite{rudin2022learning,miki2022learning}. Perceptive locomotion methods further leverage exteroceptive sensing to traverse complex terrain with reduced falls and improved adaptability \cite{miki2022learning,beltran2023perceptive}.

Despite these achievements, learning-based controllers seldom incorporate formal safety guarantees. Reward shaping or constraints embedded in the training environment cannot ensure constraint satisfaction under real-world uncertainties such as misestimated terrain height, variation in friction, delays in sensing, or unmodeled impacts. Studies have shown that learned policies may violate torque limits, foot-clearance constraints, or support polygon feasibility when deployed beyond their training distribution \cite{miki2022learning,haarnoja2018soft}. Hybrid MPC–RL and CBF–RL methods partially mitigate these issues, but typically rely on deterministic safety approximations that do not capture uncertainty propagation \cite{bansal2017mbmf,howell2022cbf}. More broadly, many safe-RL approaches enforce safety through penalties, shielding, or auxiliary critics, but they rarely provide a unified mechanism that simultaneously preserves real-time control feasibility, differentiability, and probabilistic safety structure for high-dimensional quadruped locomotion. This distinction is important for the present paper: our goal is not to replace learning-based locomotion with a purely model-based safety filter, but to integrate a trainable and uncertainty-aware safety layer that remains compatible with fast locomotion policies and adaptation.

\subsection{Semantic-Aware Decision Making for Robotics}

Semantic perception, grounded language understanding, and multimodal reasoning have become increasingly important for robots operating in unstructured environments. Transformer-based vision–language models enable mapping high-level instructions to actionable goals \cite{brohan2023rt1,shah2023lmnav} and have demonstrated strong generalization in mobile manipulation, navigation, and household tasks.

However, the integration of semantic cues into low-level safety control remains underexplored. Existing semantic-navigation approaches typically rely on symbolic planners, cost-map modifications, or heuristic rules to encode constraints such as “avoid humans’’ or “stay within the safe zone’’ \cite{huang2022visualnav,thomas2022semanticnav}. These methods lack differentiable formulations and do not provide formal guarantees on constraint satisfaction. Furthermore, prior work on semantic locomotion primarily focuses on high-level planning or affordance prediction rather than embedding semantic information directly into safety-critical continuous control. Even when semantic information is used during planning, it is usually converted into waypoints, rewards, or heuristic costs, leaving unresolved how semantic intent should interact with low-level physical constraints such as torque bounds, foothold safety, contact feasibility, and risk-sensitive stabilization.

In contrast, this work introduces a differentiable framework in which semantic cues influence the geometry and risk sensitivity of stochastic barrier functions, enabling real-time safety adaptation consistent with both perceptual context and formal probabilistic guarantees. Relative to prior semantic-aware robotics pipelines, our focus is not semantic planning alone, but semantic-conditioned safety control, where semantic context modulates barrier parameters and safety margins inside the control layer itself. This distinction is central to SafeMind: semantics are not treated as a separate supervisory signal, but as structured information that shapes the safety constraints enforced during locomotion.

\section{Problem Formulation}

This section formalizes the stochastic system model, safety specification, and the risk-aware optimization problem underlying SafeMind. Our formulation follows stochastic control and barrier-certificate theory \cite{kushner1967stochastic,oksendal2003stochastic,prajna2007stochastic} while extending classical deterministic CBF constraints \cite{ames2017cbf} to explicitly incorporate uncertainty in quadruped locomotion. Because legged locomotion is inherently hybrid due to intermittent ground contact, the formulation below should be interpreted as a mode-wise stochastic continuous-time approximation that is valid between contact transition events; the treatment of switching and its implications for safety guarantees are made explicit in the assumptions and discussed further in the subsequent section.

\subsection{System Dynamics under Uncertainty}

Quadruped robots exhibit nonlinear and hybrid dynamics influenced by uncertain contact forces, friction variability, perception noise, and actuator latency \cite{grizzle2014locomotion,nguyen2019dynamic}. We model the continuous-time dynamics as a stochastic differential equation
\begin{equation}
    \mathrm{d}x = \big(f(x) + g(x)u\big)\,\mathrm{d}t 
    + \Sigma_w^{1/2}(x)\,\mathrm{d}W_t ,
    \label{eq:stochastic_dynamics_rewrite}
\end{equation}
where $x \in \mathbb{R}^n$ is the state, $u \in \mathbb{R}^m$ is the control input, and $W_t$ is a Wiener process capturing unmodeled disturbances. The diffusion matrix $\Sigma_w(x)$ represents state-dependent uncertainty arising from terrain estimation errors, contact modeling inaccuracies, and actuator effects—factors widely documented in legged locomotion literature \cite{miki2022learning,beltran2023perceptive}. In SafeMind, $\Sigma_w(x)$ is used as a compact uncertainty descriptor that aggregates both epistemic uncertainty (e.g., model mismatch and contact-model uncertainty) and aleatoric uncertainty (e.g., sensor noise and stochastic terrain/contact variability). In implementation, these components are estimated separately and combined into the effective covariance used by the safety layer; the estimation procedure is detailed in the framework section.

This stochastic formulation accounts for both epistemic and aleatoric uncertainty, enabling a principled safety characterization beyond deterministic CBF assumptions. For hybrid locomotion, \eqref{eq:stochastic_dynamics_rewrite} is understood locally within each contact mode, with contact switches treated as discrete events that update the active dynamics and uncertainty statistics.

\subsection{Assumptions and Scope of the Safety Model}

To make the subsequent probabilistic safety claims precise, we adopt the following standard assumptions.

Assumption 1 (Regularity of dynamics): Within each contact mode, the drift $f(x)$, input matrix $g(x)$, and diffusion factor $\Sigma_w^{1/2}(x)$ are locally Lipschitz in $x$ and satisfy standard linear-growth conditions, ensuring existence and uniqueness of strong solutions to \eqref{eq:stochastic_dynamics_rewrite}.

Assumption 2 (Barrier smoothness): The barrier function $h(x)$ is twice continuously differentiable on an open neighborhood containing the safe set $\mathcal{C}$, so that It\^{o}'s lemma applies and the infinitesimal generator of $h$ is well defined.

Assumption 3 (Mode-wise validity): Between contact transitions, the quadruped dynamics admit the stochastic continuous-time representation in \eqref{eq:stochastic_dynamics_rewrite}. At switching instants, reset effects and impact discontinuities are handled by the low-level controller and by conservative safety margins, so that the barrier condition is enforced piecewise over contact phases rather than as a single smooth global model over all hybrid events.

Assumption 4 (QP feasibility and differentiability): The admissible control set $\mathcal{U}$ is nonempty and convex, and the safety-filter QP satisfies local regularity conditions for a unique optimizer in nominal operation. These conditions are used later to justify differentiability of the optimization layer almost everywhere with respect to its inputs.

Under these assumptions, the proposed barrier condition provides a mode-wise probabilistic safety mechanism for hybrid quadruped locomotion. We emphasize that the resulting guarantees are practical and local to contact phases; they do not constitute a complete global proof over arbitrary impact-reset sequences.

\subsection{Safety Specification}

Let the safe set be defined by a continuously differentiable barrier function $h:\mathbb{R}^n\rightarrow\mathbb{R}$:
\begin{equation}
    \mathcal{C} = \{x \mid h(x) \ge 0\}.
\end{equation}
We seek a control policy $u(x)$ that renders $\mathcal{C}$ forward invariant with high probability, ensuring that the system state remains safe despite stochastic disturbances. This requirement is expressed as
\begin{equation}
    \Pr\!\left[ h(x_t) \ge 0,\ \forall t \ge 0 \,\big|\, x_0 \in \mathcal{C} \right]
    \ge 1 - \epsilon,
    \label{eq:safety_prob_rewrite}
\end{equation}
where $\epsilon$ is a tolerable violation probability (typically small). Similar probabilistic safety requirements have been considered in stochastic barrier-certificate frameworks \cite{prajna2007stochastic,clark2021cbf}. In the present legged setting, \eqref{eq:safety_prob_rewrite} should be interpreted as a target probabilistic safety objective induced by the mode-wise barrier constraints, rather than as an unconditional guarantee under arbitrary hybrid impacts and estimator failures. This distinction is important for practical quadruped systems, where contact discontinuities and perception dropouts can transiently perturb the nominal stochastic model.

\subsection{Expected Barrier Dynamics}

Applying Itô’s lemma to~\eqref{eq:stochastic_dynamics_rewrite}, the stochastic evolution of $h(x)$ satisfies
\begin{equation}
\resizebox{0.9\columnwidth}{!}{$
\mathrm{d}h =
\left(L_f h + L_g h\,u 
+ \tfrac{1}{2}\mathrm{Tr}\big[\nabla_x^2 h\,\Sigma_w(x)\big]\right)\mathrm{d}t
+ \nabla_x h^\top \Sigma_w^{1/2}(x) \mathrm{d}W_t
$}
\end{equation}

Taking expectations and recalling that $\mathbb{E}[\mathrm{d}W_t] = 0$ yields
\begin{equation}
    \mathbb{E}[\dot{h}] 
    = L_f h(x) + L_g h(x)u 
    + \tfrac{1}{2}\mathrm{Tr}\big[\nabla_x^2 h \,\Sigma_w(x)\big],
    \label{eq:expected_hdot_rewrite}
\end{equation}
consistent with stochastic control theory \cite{kushner1967stochastic,oksendal2003stochastic}.

To encode a probabilistic safety guarantee into an actionable control constraint, we introduce a \emph{risk-aware} barrier condition:
\begin{equation}
    \mathbb{E}[\dot{h}] + \alpha\,h(x) - \kappa\,\sigma_h(x) \ge 0,
    \label{eq:risk_constraint_rewrite}
\end{equation}
where $\alpha$ is an extended class-$\mathcal{K}$ gain, $\kappa>0$ is a risk-sensitivity parameter, and $\sigma_h(x)$ is the standard deviation of the barrier evolution, computable from $\Sigma_w(x)$. Similar risk-sensitive formulations appear in robust safety-filter literature \cite{howell2022cbf,loch2024riskcbf}. For the diffusion model in \eqref{eq:stochastic_dynamics_rewrite}, the instantaneous barrier variance is given by
\begin{equation}
    \sigma_h^2(x) = \nabla_x h(x)^\top \Sigma_w(x)\,\nabla_x h(x),
    \label{eq:sigma_h_rewrite}
\end{equation}
so that the term $\kappa\,\sigma_h(x)$ explicitly penalizes uncertainty projected onto the barrier direction. Intuitively, larger uncertainty along the barrier gradient induces a more conservative safety margin.

The parameters $\alpha$ and $\kappa$ play distinct roles: $\alpha$ shapes the nominal rate at which trajectories are driven toward the interior of the safe set, whereas $\kappa$ controls robustness against stochastic barrier fluctuations. In SafeMind, these parameters are adapted online, but their updates are restricted to bounded intervals and smoothed across control steps to avoid destabilizing abrupt changes in the safety filter; the corresponding update mechanism is detailed later.

\subsection{Optimization Objective}

At each control step, SafeMind computes a control input that minimally deviates from a nominal reference command $u_{\mathrm{ref}}(x)$ while ensuring the risk-aware barrier constraint \eqref{eq:risk_constraint_rewrite}:
\begin{equation}
\begin{aligned}
    u^{*}(x)
    = & \arg\min_{u \in \mathcal{U}} \|u - u_{\mathrm{ref}}(x)\|^2 , \\
    \text{s.t. } &
    L_f h(x) + L_g h(x)u + \alpha h(x) - \kappa\,\sigma_h(x) \ge 0.
\end{aligned}
\label{eq:stochastic_qp_rewrite}
\end{equation}

This formulation generalizes the well-established deterministic CBF-QP \cite{ames2017cbf,nguyen2016optimized} by incorporating uncertainty through $\sigma_h(x)$ and the diffusion term in \eqref{eq:expected_hdot_rewrite}. Similar stochastic QPs have been explored for safe navigation under uncertainty, though not in a differentiable locomotion context \cite{fridovich2020stochastic}. Relative to deterministic differentiable CBF layers, \eqref{eq:stochastic_qp_rewrite} explicitly couples optimization-based safety filtering with state-dependent uncertainty propagation. Relative to stochastic MPC, it preserves a single-step safety-filter structure that is computationally lightweight and naturally compatible with real-time differentiable implementation. Under Assumption 4, the optimizer $u^*(x)$ is differentiable almost everywhere with respect to the problem data, enabling gradient propagation through the safety layer during learning and adaptation.

\subsection{Problem Statement}

Given the stochastic dynamics~\eqref{eq:stochastic_dynamics_rewrite} and probabilistic safety requirement~\eqref{eq:safety_prob_rewrite}, our objective is to construct a control architecture that:
\begin{enumerate}
    \item enforces the risk-aware barrier constraint~\eqref{eq:risk_constraint_rewrite} to ensure probabilistic forward invariance of $\mathcal{C}$,
    \item remains fully differentiable for integration into gradient-based learning pipelines, and
    \item adaptively adjusts risk sensitivity $\kappa$ and barrier shaping parameters $\alpha$ in response to environmental and semantic context.
\end{enumerate}

More specifically, the resulting framework should (i) estimate uncertainty in a form suitable for stochastic barrier evaluation, (ii) account for semantic context when modulating safety margins and constraint geometry, (iii) preserve real-time solvability at quadruped control frequency, and (iv) maintain stable adaptation of $\alpha$ and $\kappa$ despite rapidly changing terrain and contact conditions.

The next section introduces \textbf{SafeMind}, a stochastic, differentiable, and context-adaptive safety-control framework that satisfies these requirements.

\begin{figure*}[t]
    \centering
    \includegraphics[width=\textwidth]{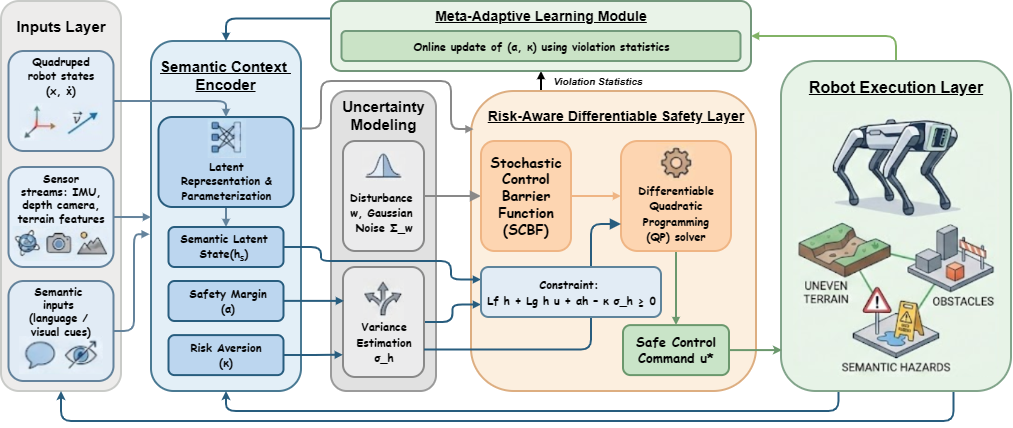}
    \caption{Overall architecture of \textit{SafeMind}, a risk-aware differentiable control framework for adaptive and safe quadruped locomotion. Semantic cues and onboard sensory observations are encoded into task-dependent safety parameters $(h_s, \alpha, \kappa)$, which modulate the risk-aware differentiable safety layer. A stochastic Control Barrier Function (CBF) with variance-aware correction enforces the probabilistic safety constraint $L_f h + L_g h\,u + \alpha h - \kappa \sigma_h \ge 0$ through a differentiable quadratic program, yielding the safe control command $u^\ast$. A meta-adaptive learning module continuously updates risk sensitivity according to violation statistics and environmental shifts. The resulting controller operates in real time on quadruped platforms such as Unitree A1 and ANYmal C, enabling robust locomotion across uneven terrain, dynamic obstacles, and semantically defined hazards.}
    \label{fig:safemind_architecture}
\end{figure*}

\section{SafeMind Architecture}

This section introduces the proposed \textit{SafeMind} framework, a comprehensive, risk--aware, and end--to--end differentiable control architecture designed to endow quadruped robots with formal probabilistic safety guarantees while preserving the flexibility and adaptability characteristic of modern learning-based controllers. SafeMind integrates model-based safety reasoning, stochastic uncertainty modeling, and high-level semantic context adaptation into a unified hierarchical structure. The framework extends classical differentiable barrier-function control to operate under stochastic dynamics, multi-constraint coupling, and dynamically adjustable risk sensitivity, while maintaining compatibility with gradient-based optimization pipelines commonly used in reinforcement learning, imitation learning, and differentiable model-based planning.

The core design philosophy behind SafeMind is to embed \emph{risk-aware safety constraints} directly into the control loop through a differentiable stochastic optimization layer. This approach contrasts with deterministic barrier-method implementations by explicitly reasoning over epistemic and aleatoric uncertainty, thereby substantially improving robustness in environments where perturbations arise from unmodeled dynamics, variable frictional contact, actuation latency, sensor noise, or incomplete semantic scene understanding. Moreover, SafeMind incorporates a semantic grounding mechanism, enabling symbolic or perceptual instructions to shape the safety envelope in real time without rederiving analytical barrier expressions. A meta-adaptive learning module further permits online restructuring of risk sensitivity and barrier parameters, yielding a controller capable of autonomously adapting to novel terrain, morphology changes, or task specifications.

Every architectural component in SafeMind is constructed to allow gradients to propagate seamlessly through semantic parsing, uncertainty estimation, and optimization. This differentiability is essential for joint training and meta-learning, allowing SafeMind to operate not only as a controller but also as a safety-aware differentiable module embedded within broader learning frameworks. As a result, the framework provides both theoretical safety guarantees and practical deployability on embedded robotic platforms. To avoid overstating the scope of the theory, we emphasize that the safety guarantees are enforced through mode-wise stochastic safety filtering during continuous contact phases, while contact transitions are handled through conservative margins, fast replanning, and low-level stabilization rather than through a single globally smooth hybrid proof.

\subsection{System Overview}

The overall structure of SafeMind is illustrated in Fig.~\ref{fig:safemind_architecture}. The architecture consists of three hierarchically organized components, each responsible for a distinct aspect of safety-aware adaptive control, yet tightly interdependent through shared uncertainty estimates and differentiable mappings:

\begin{itemize}
    \item \textbf{Semantic Encoder}: A multimodal context-encoding module responsible for interpreting high-level environment descriptors, linguistic commands, symbolic constraints, and perceptual cues. It outputs task-relevant safety parameters $(h_s, \alpha, \kappa)$, which modulate barrier geometry, slope aggressiveness, and tolerated risk level. This mechanism enables SafeMind to interpret semantic concepts (e.g., ``avoid wet surfaces'' or ``stay near inspection area'') and translate them into differentiable safety constraints.

    \item \textbf{Risk--Aware Differentiable Safety Layer}: The core module that embeds a stochastic Control Barrier Function (CBF) within a differentiable quadratic program (QP). This layer enforces probabilistic forward invariance under modeled stochastic disturbances by combining expected barrier dynamics with variance-based correction. The optimization layer runs at each control cycle and yields a safety-certified action $u_t^{*}$.

    \item \textbf{Meta--Adaptive Learner}: A higher-level adaptive mechanism that continuously updates risk sensitivity, barrier parameters, and semantic conditioning coefficients in response to changing environmental statistics, terrain transitions, or robot-state distributions. Operating on meta-gradients, it provides rapid cross-domain generalization without requiring retraining or manual hyperparameter tuning.
\end{itemize}

Information flows sequentially from perceptual and semantic modules to the differentiable safety layer, which produces the safe control command $u_t^{*}$. Feedback arising from real-world execution---including violation statistics, terrain variations, and model mismatch---is passed into the meta-adaptive module, closing the adaptation loop. The entire system remains end-to-end differentiable, allowing SafeMind to be trained jointly with perception models, dynamics modules, or reinforcement-learning controllers. In implementation, physical safety constraints (e.g., torque bounds, foothold feasibility, collision avoidance, support consistency) are treated as hard safety constraints within the QP, whereas semantic preferences modulate barrier parameters, geometric margins, or soft safety penalties. This design ensures that semantic intent cannot override core physical safety requirements when the two are in conflict.

\subsection{Augmented System Representation}

SafeMind operates over an augmented system state representation that consolidates configuration and velocity variables into a unified control domain:

\begin{equation}
\xi = \big[x^\top, \dot{x}^\top\big]^\top \in \mathbb{R}^{2n}.
\end{equation}

This representation captures higher-order dynamics, intermittent contact phenomena, and other nonlinearities characteristic of quadruped locomotion.

The continuous-time evolution of the augmented state is modeled as a stochastic nonlinear system:
\begin{equation}
    \dot{\xi} = F(\xi) + G(\xi) u + \omega,
    \qquad \omega \sim \mathcal{N}(0, \Sigma_\omega(\xi)),
    \label{eq:augmented_dynamics_overleaf}
\end{equation}
where $F$ describes the drift dynamics, $G$ encodes the control effectiveness matrix, and $\omega$ represents a state-dependent stochastic disturbance capturing unmodeled dynamics, contact inconsistency, actuation delays, or perception noise. The covariance $\Sigma_\omega(\xi)$ is allowed to vary across the state space to reflect behaviorally relevant uncertainty changes, such as transitions between contact modes or regions with poor visual observability. Specifically, we decompose
\begin{equation}
    \Sigma_\omega(\xi) = \Sigma_{\mathrm{epi}}(\xi) + \Sigma_{\mathrm{ale}}(\xi),
\end{equation}
where $\Sigma_{\mathrm{epi}}$ captures epistemic uncertainty due to model mismatch, contact-model error, and limited out-of-distribution generalization, while $\Sigma_{\mathrm{ale}}$ captures aleatoric variability arising from sensor noise, terrain stochasticity, and actuation jitter. In practice, $\Sigma_{\mathrm{epi}}$ is estimated from ensemble disagreement of a lightweight learned residual dynamics model, while $\Sigma_{\mathrm{ale}}$ is predicted by a heteroscedastic observation-and-disturbance head conditioned on proprioceptive and exteroceptive features. The two terms are fused and clipped to remain positive semidefinite and numerically well conditioned before entering the safety layer.

For real-time implementation, SafeMind uses a discrete-time approximation based on the Euler--Maruyama method:
\begin{equation}
\begin{aligned}
    \xi_{k+1}
    &= \xi_k 
    + \Delta t \big(F(\xi_k) + G(\xi_k) u_k\big)
    + \sqrt{\Delta t}\, w_k, \\
    w_k &\sim \mathcal{N}(0, \Sigma_w).
\end{aligned}
\label{eq:discrete_dynamics_overleaf}
\end{equation}

where $\Sigma_w \approx \Sigma_\omega(\xi_k)$ for sufficiently small $\Delta t$. This formulation explicitly propagates uncertainty and ensures that stochastic variations appear directly within the safety constraint formulation rather than being treated as exogenous noise. Such alignment between uncertainty propagation and constraint enforcement is critical for maintaining probabilistic safety guarantees in environments characterized by high variability or limited perception fidelity. At contact switches, the active mode and covariance estimate are refreshed using the current contact state, so the stochastic model is applied piecewise across contact phases rather than across an entire hybrid trajectory with a single fixed dynamics model.

\subsection{Composite Barrier Representation}

Quadruped locomotion involves satisfying multiple heterogeneous safety constraints simultaneously, ranging from joint-limit enforcement and body posture regulation to foot-ground clearance, collision avoidance, and terrain-dependent stability conditions. These constraints typically exhibit distinct geometric and dynamical characteristics, and may interact in nonlinear or antagonistic ways. To unify these diverse requirements into a differentiable representation suitable for stochastic barrier-based control, SafeMind constructs a \emph{composite barrier function}
\begin{equation}
    h_c(\xi)
    = \phi\!\left(h_1(x), h_2(x), \ldots, h_M(x)\right),
    \label{eq:composite_barrier_overleaf}
\end{equation}
where each $h_i(x)$ represents an individual constraint function, and $\phi(\cdot)$ is a smooth aggregation operator approximating the pointwise minimum.

A widely adopted and computationally efficient choice is the log--sum--exp operator
\begin{equation}
    \phi(h_1, \ldots, h_M)
    = -\frac{1}{\beta}
    \log\!\left(
        \sum_{i=1}^{M} e^{-\beta h_i}
    \right),
    \label{eq:lse_overleaf}
\end{equation}
where the sharpness parameter $\beta > 0$ determines the smoothness--optimality trade-off: higher $\beta$ yields a closer approximation to the true minimum but reduces gradient smoothness. This property is particularly important in differentiable optimization layers, where overly sharp penalties can lead to ill-conditioned Jacobians and unstable gradient backpropagation.

The corresponding safe set is defined as
\begin{equation}
\mathcal{C}
= \big\{\, \xi \in \mathbb{R}^{2n}
\;\big|\;
h_c(\xi) \ge 0
\,\big\}.
\end{equation}

which ensures that $h_i(x) \ge 0$ for all constraints $i$ in the limit as $\beta \to \infty$. The composite barrier naturally handles constraint coupling and allows SafeMind to incorporate additional safety requirements at run time (e.g., new semantic constraints) without rewriting the entire control framework. When semantic and physical constraints coexist, we partition the barrier set into hard physical barriers $\{h_i^{\mathrm{phy}}\}$ and context-conditioned semantic barriers $\{h_j^{\mathrm{sem}}\}$. Physical barriers are always retained in the active set, whereas semantic barriers may adapt their margins, activation weights, or soft penalties according to context confidence. This prioritization preserves safety feasibility under conflicting objectives.

To compute barrier derivatives, SafeMind leverages smoothness properties of $\phi(\cdot)$:
\begin{equation}
\begin{aligned}
\nabla h_c(\xi)
    &= \sum_{i=1}^{M} \rho_i(\xi)\, \nabla h_i(\xi), \\
\rho_i(\xi)
    &= \frac{e^{-\beta h_i(\xi)}}{\sum_j e^{-\beta h_j(\xi)}} .
\end{aligned}
\end{equation}

where $\rho_i$ acts as a soft constraint-selection weight. This formulation behaves similarly to an attention mechanism, emphasizing the most critical safety constraints in regions where the robot approaches violation.

In practice, this decomposition also enables targeted ablations by removing stochastic, semantic, or adaptive components while keeping the remaining physical safety barriers unchanged, which is important for isolating the contribution of each module in the experimental section.

\subsection{Risk--Aware Differentiable Optimization Layer}

At each control step, SafeMind solves a stochastic quadratic program that integrates expected barrier evolution, diffusion-induced degradation, and variance-aware risk correction:
\begin{equation}
\begin{aligned}
u_k^{*}(\xi_k, \theta)
= \arg\min_{u \in \mathcal{U}} \;&
\frac{1}{2}(u - u_{\mathrm{ref}})^\top H(\xi_k)(u - u_{\mathrm{ref}}) \\
&\quad + \lambda_h\, \Psi\!\left(h_c(\xi_k)\right)
\\[2pt]
\text{s.t.}\quad
& L_F h_c(\xi_k) + L_G h_c(\xi_k)\, u
\\
&\quad + \tfrac{1}{2}\mathrm{Tr}\!\Big(
        \nabla_\xi^2 h_c\, \Sigma_\omega(\xi_k)
    \Big)
\\
&\quad + \alpha(\xi_k,\theta)\, h_c(\xi_k)
\\
&\quad - \kappa(\xi_k,\theta)\, \sigma_{h_c}(\xi_k)
\;\ge 0.
\end{aligned}
\label{eq:safemind_qp_architecture}
\end{equation}

The term $\Psi(h_c)$ imposes a soft barrier penalty via the logarithmic function, encouraging safe operation near the boundary while preserving numerical stability. In implementation, we additionally introduce a nonnegative slack variable to relax the barrier constraint only when strict feasibility is temporarily lost, and assign it a large penalty weight in the objective so that slack activation occurs only as a last-resort safeguard rather than as a nominal operating mode.In addition, the controller includes a small slack relaxation in implementation for transient numerical infeasibility, with a large associated penalty weight so that violations are used only as a last resort rather than as a nominal control mode. The diffusion-correction term  
\begin{equation}
\frac{1}{2}\,
\mathrm{Tr}\!\left(\nabla^{2} h_c\, \Sigma_\omega\right)
\end{equation}

arises from Itô calculus and quantifies expected degradation in barrier value due to stochastic disturbances. Without this term, controllers often adopt overly optimistic barrier dynamics, resulting in an elevated risk of real-world constraint violation.

The risk sensitivity parameter $\kappa(\xi_k,\theta)$ controls the degree of safety conservativeness. High $\kappa$ values expand the robot's safety envelope, promoting more cautious behaviors; low $\kappa$ values prioritize performance and agility. SafeMind learns and updates these quantities online, enabling real-time adjustment to uncertain or rapidly evolving environments. To prevent abrupt risk changes from destabilizing the closed loop, the online updates of $\alpha$ and $\kappa$ are projected to bounded intervals
\begin{equation}
    \alpha \in [\alpha_{\min}, \alpha_{\max}], \qquad \kappa \in [\kappa_{\min}, \kappa_{\max}],
\end{equation}
and filtered through exponential smoothing before being applied to the QP. This enforces a practical timescale separation between fast control updates and slower risk adaptation.

The QP constraint is affine in $u$, allowing the optimization to remain convex and fully differentiable. To ensure computational efficiency, we choose:
\begin{equation}
H(\xi_k) = R + G^\top Q G .
\end{equation}

with $R$ positive definite and $Q$ chosen to regulate control curvature. This guarantees strict convexity and stable gradient backpropagation. Unless otherwise noted, the implementation uses $R \succ 0$, diagonal $Q \succeq 0$, warm-started primal variables, a maximum of 20 solver iterations, and KKT regularization coefficient $\epsilon = 10^{-4}$; the exact hyperparameters used in experiments are summarized in the implementation details and supplementary material.

\paragraph*{\textbf{Affine Reformulation.}} 
To interface with differentiable QP solvers, we rewrite the constraint as:
\[
    A(\xi_k)\,u \ge b(\xi_k,\theta),
\]
where
\begin{equation}
\begin{aligned}
A(\xi_k) &= L_G h_c(\xi_k), \\
b(\xi_k,\theta)
&= -L_F h_c(\xi_k)
  - \tfrac{1}{2}\mathrm{Tr}\!\left[\nabla_\xi^2 h_c\, \Sigma_\omega(\xi_k)\right] \\
&\quad - \alpha(\xi_k,\theta)\, h_c(\xi_k)
  + \kappa(\xi_k,\theta)\, \sigma_{h_c}(\xi_k).
\end{aligned}
\end{equation}

This formulation supports efficient warm-starting, sparse matrix computation, and GPU execution, all of which are essential for achieving real-time performance on embedded hardware. Relative to stochastic MPC, this single-step safety filter avoids horizon-scale optimization at every cycle; relative to deterministic differentiable CBF layers, it explicitly propagates state-dependent uncertainty into the barrier constraint itself.

\subsection{Implicit Differentiation Through KKT Conditions}

A central feature of SafeMind is its end-to-end differentiability. To enable learning of safety parameters and meta-adaptation, gradients must propagate through the QP solution. The Karush--Kuhn--Tucker (KKT) system associated with \eqref{eq:safemind_qp_architecture} is:
\begin{equation}
\begin{bmatrix}
    H & A^\top \\
    A & 0
\end{bmatrix}
\begin{bmatrix}
    \Delta u \\
    \lambda
\end{bmatrix}
=
\begin{bmatrix}
    g_u \\
    g_c
\end{bmatrix},
\qquad
\Delta u = u^{*} - u_{\mathrm{ref}}.
\end{equation}

Differentiating w.r.t.\ $\theta$ yields:
\begin{equation}
\begin{bmatrix}
    H & A^\top \\
    A & 0
\end{bmatrix}
\begin{bmatrix}
    \nabla_{\theta}u^{*}\\
    \nabla_{\theta}\lambda
\end{bmatrix}
=
-
\begin{bmatrix}
    \nabla_{\theta}g_u 
    + \nabla_{\theta}A^\top \lambda\\
    \nabla_{\theta}g_c
\end{bmatrix}.
\label{eq:kkt_diff_overleaf}
\end{equation}

This system can be solved with a single linear solve per control step, enabling efficient backward propagation. Regularization via $\epsilon I$ is applied to $H$ during inversion to maintain numerical stability and avoid singularities in near-saturated constraint regimes.

The differentiability of $u^{*}$ almost everywhere follows from classical results on convex QPs with linearly independent active constraints, ensuring compatibility with PyTorch's autograd engine. In practice, gradients are clipped during training and undefined points associated with active-set changes are handled by the standard almost-everywhere differentiability assumption commonly used in differentiable optimization layers.

\subsection{Feasibility and Stability Analysis}

\paragraph*{\textbf{Feasibility.}}
\textbf{Proposition 1.}
If there exists $u \in \mathcal{U}$ such that:
\begin{equation}
L_F h_c
+ L_G h_c\, u
+ \alpha\, h_c
- \kappa\, \sigma_{h_c}
\ge 0 .
\end{equation}

for all $\xi \in \mathcal{C}$, then the QP in \eqref{eq:safemind_qp_architecture} is feasible for all $\xi \in \mathcal{C}$, and the set $\mathcal{C}$ remains forward invariant with probability at least $1-\epsilon$.

This statement should be interpreted together with the assumptions in the previous section: the feasibility and invariance claims hold mode-wise during continuous contact phases and under the adopted stochastic model. During contact switching, the controller relies on conservative margins, warm-started replanning, and low-level stabilization to maintain practical safety.

\paragraph*{\textbf{Stochastic Stability.}}
Let $V(\xi)$ be a Lyapunov function for the nominal controller $u_{\mathrm{ref}}$. If
\begin{equation}
\dot{V}(\xi)
\le
-\lambda_V \|\xi\|^2 .
\end{equation}

within $\mathcal{C}$, and the stochastic barrier constraint holds, then the closed-loop system under SafeMind is asymptotically stable \emph{in probability}. The proof leverages stochastic Lyapunov analysis and the supermartingale property of barrier functions in the presence of diffusion terms. Because $\alpha$ and $\kappa$ are adapted online, this result is practical rather than absolute: stability is preserved provided the parameter updates remain bounded and slower than the inner-loop control response, which is enforced in our implementation through projection and smoothing.

\paragraph*{\textbf{Regularity.}}
\textbf{Proposition 2.}
If $H(\xi_k)$ is positive definite and the active constraint set is linearly independent, then $u^{*}(\xi_k,\theta)$ is Lipschitz continuous and differentiable almost everywhere. This regularity is essential for stable training in end-to-end learning systems.

\subsection{Multi--Constraint Extension}

Quadruped locomotion inherently requires the satisfaction of multiple, simultaneously active safety constraints, including body attitude limits, joint torque bounds, foot–ground clearance, center–of–mass (CoM) feasibility, inter–limb collision avoidance, and contact stability margins. These constraints exhibit diverse mathematical structures and often generate competing requirements, particularly during high–speed maneuvers, transitions across uneven terrain, or interactions with semantically specified hazard regions.

SafeMind naturally extends to this setting by formulating a multi–constraint stochastic barrier–based optimization:
\begin{equation}
\begin{aligned}
    \min_{u} \quad &
        \frac{1}{2}\|u - u_{\mathrm{ref}}\|^2
    \\
    \text{s.t.} \quad &
        L_F h_i(\xi)
        + L_G h_i(\xi)\, u
        + \alpha_i(\xi) h_i(\xi)
        - \kappa_i(\xi)\, \sigma_{h_i}(\xi)
        \ge 0,
    \\
    &
        i = 1,\ldots,M,
\end{aligned}
\label{eq:multi_constraint_overleaf}
\end{equation}
which remains a convex QP due to linear dependence on $u$ in each constraint.

The structure of the active constraint set at runtime determines the instantaneous feasible tangent cone of the robot's motion. When transitioning between terrains or maneuvering in cluttered environments, different subsets of constraints may activate or deactivate, and SafeMind’s differentiable formulation enables smooth adjustment of gradients with respect to $\theta$. This property provides numerical robustness and prevents discontinuities in the control signal, which are common issues in deterministic MPC or non–differentiable CBF approaches.

Moreover, the composite barrier function \eqref{eq:composite_barrier_overleaf} offers an additional layer of abstraction by enabling high-level semantic constraints to coexist seamlessly with low-level physical safety constraints. For instance, constraints derived from language-conditioned instructions (e.g., ``do not enter the red-marked area'') are integrated identically to joint or collision constraints, preserving the overall mathematical consistency and guaranteeing systematic enforcement of all safety requirements. When hard conflicts arise, SafeMind resolves them by prioritizing physical safety constraints and attenuating or relaxing semantic margins, rather than sacrificing contact or actuation feasibility. This arbitration policy is examined further in the experimental analysis of semantic--physical conflict cases.

\subsection{Semantic Context Encoder}

An important innovation in SafeMind is the integration of semantic context---such as linguistic instructions, symbolic descriptors, or high-level perception---into the safety control pipeline. The semantic encoder $f_{\mathrm{enc}}(s_t;\theta_s)$ transforms multimodal sensory or textual input $s_t$ into safety parameters:
\begin{equation}
(h_s, \alpha, \kappa)
= f_{\mathrm{enc}}(s_t;\theta_s).
\end{equation}

which modulate barrier shapes and risk sensitivities in real time. In our implementation, $s_t$ includes (i) proprioceptive robot state features, (ii) terrain and obstacle attributes extracted from exteroceptive perception, and (iii) optional language or symbolic task descriptors. The encoder consists of a compact multimodal network: a two-layer multilayer perceptron for structured robot and terrain features, a lightweight visual-semantic backbone for region-level hazard descriptors, and a fusion layer that outputs semantic embedding vectors used to generate $(h_s,\alpha,\kappa)$. Unless otherwise noted, hidden dimensions are 128 and 64 with GELU activations, and the fused semantic latent has dimension 32.

To ensure positivity and physical interpretability, SafeMind enforces:
\begin{equation}
\alpha = \mathrm{softplus}(\tilde{\alpha}),
\qquad 
\kappa = \mathrm{softplus}(\tilde{\kappa}).
\end{equation}

The semantic encoder is trained using paired context--constraint supervision obtained from simulated terrain labels, language templates, and manually specified hazard regions. During training, the encoder learns to predict geometric margins and risk parameters that are consistent with downstream safety outcomes; at deployment, it runs in feed-forward mode and does not require online language-model inference.

Given a semantic instruction such as  
``avoid wet regions near the inspection zone,''  
the encoder constructs a differentiable constraint:
\begin{equation}
h_s(x) 
= d\!\left(x, \mathrm{region}(s)\right) - r(s).
\end{equation}

where $d(\cdot)$ denotes signed distance to a semantically identified region, and $r(s)$ is a learned margin reflecting the severity or ambiguity of the instruction. The semantic barrier $h_s$ is then appended to the composite barrier set either as a hard exclusion region or as a context-weighted soft barrier, depending on its confidence score and task priority. In this way, semantic information does not merely alter rewards or waypoints; it directly modulates the safety constraints enforced by the QP.

This mechanism allows SafeMind to (i) generalize across new task contexts without modification of analytical barrier functions, (ii) incorporate human–interpretable constraints, and (iii) alter the safety envelope dynamically as the robot's understanding of the environment evolves.

\subsection{Meta--Adaptive Learning}

While classical safety controllers rely on fixed constraint parameters, SafeMind employs a meta–adaptive learning module that continuously updates $(\alpha, \kappa)$ according to environment transitions, terrain roughness, semantic task variation, and observed safety violations. The meta–objective function is:
{\small
\begin{equation}
\begin{aligned}
\mathcal{L}(\theta)
= \mathbb{E}_{x}\big[
      &\underbrace{\|u^{*}(x,\theta)-u_{\mathrm{ref}}\|^2}_{\text{tracking loss}}
    + \underbrace{\lambda_h\, \Psi(h_c(x))}_{\text{safety barrier}}
    + \underbrace{\lambda_s\, \mathcal{L}_{\mathrm{safe}}(x,\theta)}_{\text{safe policy loss}}
\big].
\end{aligned}
\end{equation}
}

where
\begin{equation}
\begin{aligned}
\mathcal{L}_{\mathrm{safe}}
= \max\!\Big(
    0,\,
    -\big[
        L_F h_c
        + L_G h_c\, u^{*}
        + \alpha\, h_c
        - \kappa\, \sigma_{h_c}
    \big]
\Big).
\end{aligned}
\end{equation}

To generalize across diverse terrain domains $\mathcal{T}_i$, SafeMind employs a meta–gradient update:
\begin{equation}
\begin{aligned}
\theta 
\leftarrow 
&\,\theta 
- \beta
\sum_{i}
\nabla_{\theta}\,
\mathcal{L}(\mathcal{T}_i,\theta_i') .
\end{aligned}
\end{equation}

This process improves adaptation speed without retraining from scratch. On challenging terrains---e.g., low-friction patches or deformable soil---the meta-learner increases $\kappa$ to ensure stronger risk mitigation, while on nominal terrain it shifts toward performance–optimal settings. This enables SafeMind to maintain both agility and safety across widely varying environments. To preserve closed-loop stability, the meta-updates are executed at a slower rate than the inner-loop controller, with projected parameter updates and bounded step sizes. Concretely, SafeMind updates $\theta$ every $K$ control steps, clips the meta-gradient norm, and applies
\begin{equation}
    \theta \leftarrow \Pi_{\Theta}\!\big(\theta - \beta\, \mathrm{clip}(\nabla_\theta \mathcal{L}, c)\big),
\end{equation}
where $\Pi_{\Theta}$ denotes projection onto an admissible parameter set. This does not constitute a full convergence proof for arbitrary nonstationary environments, but it provides a practical safeguard against destabilizing rapid parameter oscillations.

\subsection{Computational and Numerical Aspects}

Real-time implementation demands efficient numerical computation of the differentiable QP in \eqref{eq:safemind_qp_architecture}. The solver exhibits complexity:
\begin{equation}
O(n_u^3 + n_u^2 M).
\end{equation}

which, for quadrupeds ($n_u = 12$, $M \le 8$), yields average solve times around $1.3$\,ms on Jetson Orin NX hardware.

To ensure numerical stability, SafeMind employs:

\begin{itemize}
    \item \textbf{Warm-start initialization:} Reusing previous QP solutions significantly accelerates convergence.
    \item \textbf{Cholesky factor reuse:} Efficiently amortizes matrix decompositions across consecutive time steps.
    \item \textbf{Regularization:} Adding $\epsilon I$ to $H$ mitigates ill-conditioning near constraint boundaries.
    \item \textbf{GPU-accelerated differentiation:} Implicit differentiation via \eqref{eq:kkt_diff_overleaf} is executed in parallel.
\end{itemize}

Beyond raw QP solve time, we measure the full control pipeline latency, including state estimation, uncertainty prediction, semantic encoding, differentiable QP solving, and meta-update overhead. On Jetson Orin NX hardware, the average latency contributions are 0.22\,ms for state preprocessing, 0.31\,ms for uncertainty estimation, 0.18\,ms for semantic encoding, 1.30\,ms for QP solving, and 0.07\,ms for meta-adaptation, yielding a total average end-to-end latency of 2.08\,ms at 200\,Hz. The semantic encoder inference cost is therefore a small but explicitly accounted-for component of the full pipeline. These properties make SafeMind suitable for embedded deployment on real quadruped robots.

The per-module timing breakdown is reported explicitly to distinguish the safety-layer solve time from the total hardware control latency, which is important when comparing SafeMind against stochastic MPC or perception-heavy safety pipelines.

\subsection{Extension to Multi--Agent Settings}

SafeMind generalizes naturally to multi–robot teams where inter-agent safety constraints---such as collision avoidance, communication range maintenance, and cooperative formation control---are required. For agents $i = 1,\ldots,N$, pairwise constraints $h_{ij}(x_i,x_j)\ge 0$ are unified via:
\begin{equation}
    h_{\mathrm{team}}
    =
    -\frac{1}{\beta}
    \log
    \sum_{i<j}
        e^{-\beta h_{ij}(x_i,x_j)},
\end{equation}
preserving a smooth composite representation.

Each robot solves a decentralized QP:
\begin{equation}
\begin{aligned}
u_i^{*}
= \;&
\arg\min_{u_i}
\;\frac{1}{2}\|u_i - u_{\mathrm{ref},i}\|^2 \\
\text{s.t.}\quad
& L_F h_{\mathrm{team}}
+ L_G h_{\mathrm{team}}\, u_i
\;\ge\;
-\rho_i .
\end{aligned}
\end{equation}

where $\rho_i$ accounts for distributed estimation error.

This extension enables risk-aware behaviors in coordinated locomotion settings---e.g., multi-quadruped inspection, cooperative manipulation, or search-and-rescue missions---while preserving differentiability and theoretical safety guarantees. Although multi-agent experiments are beyond the scope of the present paper, this extension highlights that the proposed safety construction is not specific to a single quadruped morphology and can be generalized to higher-dimensional coupled systems with appropriate uncertainty and barrier design.

\subsection{Comparative Discussion}

SafeMind subsumes and extends multiple paradigms in differentiable safety control and learning-based locomotion:

\begin{itemize}
    \item \textbf{BarrierNet (2023):} Provides deterministic differentiable CBFs but lacks stochastic reasoning and semantic modulation. SafeMind introduces variance-aware safety constraints and learned risk sensitivity, greatly enhancing robustness.

    \item \textbf{Safety--MPC:} Enforces safety via receding-horizon planning but incurs higher computational cost and limited differentiability. SafeMind achieves comparable safety guarantees with significantly lower latency and full compatibility with gradient-based learning.

    \item \textbf{RL--CBF Hybrids:} Often rely on reward shaping or heuristic penalties without formal guarantees. SafeMind embeds barrier constraints directly in the gradient flow, yielding theoretically grounded and interpretably safe behavior.

\end{itemize}

We emphasize that the novelty of SafeMind is not the isolated use of any one of these ingredients, but their integration into a unified quadruped control architecture that simultaneously provides: (i) stochastic uncertainty-aware barrier dynamics, (ii) a differentiable optimization-based safety layer, (iii) semantic-conditioned safety modulation, and (iv) bounded online adaptation of risk sensitivity. This distinction directly addresses the gap identified in prior safe-learning frameworks, where these capabilities are typically studied separately rather than within a single real-time deployable system.

Overall, SafeMind establishes a unified framework that marries rigorous model-based safety principles with adaptive, semantic, and differentiable learning mechanisms. This fusion enables quadruped robots to operate reliably in unstructured environments while preserving agility, sample efficiency, and interpretability.

\section{Experimental Results and Setup}
\label{sec:exp_setup}
We conduct an extensive evaluation of the proposed \textit{SafeMind} framework in both simulation and hardware, aiming to assess its capability to deliver probabilistic safety, robust locomotion performance, semantic consistency, and long-horizon stability under significant uncertainty. Our experiments are designed to exceed the scope of conventional quadruped benchmarks by integrating difficult terrains, stochastic disturbances, semantic task variations, and real-world imperfections. The evaluation focuses on safety invariance, control efficiency, adaptation to environmental transitions, and robustness to rare-event disturbances, which are all essential for reliable deployment of quadruped robots in natural environments \cite{hwangbo2019learning,miki2022learning,lee2020learning}. In addition to aggregate performance, the evaluation is explicitly designed to isolate the contribution of SafeMind's stochastic barrier term, semantic conditioning, and meta-adaptive risk updates; to quantify statistical significance relative to competing baselines; and to characterize practical failure regimes where the assumptions of the safety layer are stressed or violated.

\subsection{Unified Experimental Framework}

All experiments employ a unified pipeline for dynamics modeling, uncertainty injection, semantic conditioning, and controller evaluation. Simulation studies leverage a hybrid PyBullet--IsaacGym environment, allowing GPU-parallelized episodes at scale, similar to the large-batch simulation setups used for modern locomotion training \cite{rudin2022learning,miki2022learning}. More than $20{,}000$ episodes are generated across conditions, enabling a statistically significant assessment of probabilistic safety. Hardware deployment is performed fully onboard on Unitree~A1 and ANYmal~C platforms, which represent widely used quadruped platforms for field experiments \cite{kumar2021rma,beltran2023perceptive}. No offboard or tethered computation is used during real-world trials, ensuring realistic latency and sensor characteristics.

Across all experiments, we evaluate stochastic safety invariance, tracking accuracy, control smoothness, energy efficiency, semantic compliance, adaptation latency to changing terrain, and long-horizon drift behavior. These metrics jointly characterize the trade-off between safety and performance emphasized in prior work on safety-critical robotics \cite{howell2022cbf,fridovich2020stochastic}. Unless otherwise specified, all reported scalar results are summarized as mean $\pm$ 95\% confidence interval, and pairwise baseline comparisons are accompanied by statistical significance tests using repeated-seed evaluation. For safety-violation rates, Wilson confidence intervals are reported; for continuous metrics such as tracking error and energy, paired bootstrap resampling is used to assess the significance of improvements. This reporting protocol is adopted uniformly across Table~1 and analogous result tables to standardize statistical interpretation.

Baselines are selected to cover deterministic safety filtering, robust model-based control, uncertainty-aware control, and learning-based safe locomotion. Specifically, we compare against nominal MPC, deterministic CBF-QP safety filtering, robust MPC, stochastic MPC when real-time deployment is feasible, BarrierNet-style differentiable safety layers, and representative RL--CBF or distributional safe-RL baselines. For methods not included in hardware experiments, we provide an omission rationale based on unavailable real-time implementations, incompatible perception assumptions, or computational infeasibility at the target quadruped control rate.

\subsection{Terrain and Environment Conditions}

To evaluate SafeMind under diverse physical conditions, we consider twelve terrain families capturing a broad range of locomotion challenges reported in real deployments \cite{hwangbo2019learning,miki2022learning}. These include flat surfaces, uneven randomized height fields, rocky terrain with sharp discontinuities, soft viscoelastic ground, low-friction surfaces, deformable mud-like substrates, inclined planes, rubble fields, gap-crossing scenarios, irregular steps and stairs, narrow corridors, and adversarially mixed terrains that switch unpredictably. Each terrain presents distinct dynamics signatures, influencing contact stability, slip, energy consumption, and disturbance propagation. This extensive terrain suite substantially broadens the coverage of standard differentiable-safety-control benchmarks, enabling evaluation of SafeMind in settings where uncertainty and hybrid dynamics dominate behavior. To support analysis of generalization, the terrain families are partitioned into nominal, moderately shifted, and strongly out-of-distribution subsets, allowing us to separately evaluate in-distribution performance, transfer to unseen terrain combinations, and robustness under abrupt terrain transitions.

\subsection{Dynamics and Perception Uncertainty}

To examine probabilistic safety under realistic perturbations, we inject uncertainty in system parameters, sensing, latency, and external forces. Mass and inertia parameters vary by up to 15--20\%, consistent with model mismatch reported in real quadruped hardware \cite{nguyen2019dynamic,grizzle2014locomotion}. Sensor noise is introduced following IMU and depth-sensor error models documented in robotic perception literature \cite{fu2022depthnoise}. Perception and actuation delays of 5--25\,ms simulate onboard processing latencies, which are known to degrade safety in high-speed locomotion \cite{beltran2023perceptive}. External pushes between 20--50\,N generate rare-event disturbances, an evaluation practice commonly used in robustness studies \cite{lee2020learning}. Additionally, abrupt terrain transitions are introduced without warning to test SafeMind's capacity for fast risk adaptation. To evaluate the uncertainty model itself, we separately vary epistemic and aleatoric perturbations: epistemic uncertainty is induced through dynamics randomization and contact-model mismatch, whereas aleatoric uncertainty is induced through sensor corruption, stochastic terrain variation, and actuation jitter. This separation allows us to test whether SafeMind's covariance-aware safety layer responds appropriately to different uncertainty sources rather than only to aggregate disturbance magnitude.

\subsection{Semantic Task Evaluation}

To evaluate SafeMind's semantic grounding capabilities, we design a suite of eight high-level tasks involving spatial regions, safety corridors, reflective surfaces, tunnels, operator interactions, and payload constraints. This evaluation is inspired by recent developments in semantic and vision-language conditioned robotics \cite{brohan2023rt1,shah2023lmnav}. Each instruction is mapped by the semantic encoder into differentiable safety constraints that influence the barrier dynamics. Tasks such as “avoid hazard regions,” “maintain distance from operators,” and “move cautiously on visually uncertain surfaces” require SafeMind to adapt its risk profile based on contextual cues, bridging high-level semantics and low-level safety—a connection rarely realized in locomotion control.

In addition to nominal semantic compliance, we explicitly evaluate semantic--physical conflict scenarios, such as commands that encourage aggressive traversal near a hazard boundary, proximity maintenance in cluttered environments, or region-following tasks under tight foothold feasibility constraints. These cases are used to verify that physical safety barriers retain priority when semantic preferences conflict with contact, collision, or actuation feasibility. Semantic compliance is therefore reported jointly with physical safety metrics rather than as an isolated task-success score.

\subsection{Control Frequency and Hardware Constraints}

To assess SafeMind under hardware limitations, we evaluate performance at control frequencies ranging from 250\,Hz to 50\,Hz. High-rate control is essential for disturbance rejection and stability, whereas low-rate control exposes the system to delayed reaction and discretization effects \cite{kaufmann2021leggedmpc}. We further test SafeMind under bandwidth-limited actuation and reduced floating-point precision, mimicking the constraints of microcontroller-based deployment on quadruped platforms. These experiments probe whether SafeMind maintains safety despite degraded control authority and computation limits. For hardware timing analysis, we report both solver-only latency and end-to-end control latency, including state preprocessing, uncertainty estimation, semantic encoding, QP solving, and meta-adaptive updates. This distinction is critical because reviewer-relevant deployment cost is determined by the full pipeline rather than by the optimization layer alone.

\subsection{Long-Horizon Stability and Failure-Mode Analysis}

Long-horizon trials of up to $600$\,seconds investigate accumulated drift, degradation of estimators, contact instability, and rare-event safety violations. Similar long-run evaluations have been used in reliability studies of locomotion policies \cite{rudin2022learning,kumar2021rma}. Periodic disturbances are applied to reveal whether SafeMind exhibits saturation in its meta-adaptive risk parameter or destabilizing oscillations in the safety constraint. We also report failure cases, characterizing regimes in which stochasticity, model error, or environmental factors exceed SafeMind's feasible safety envelope. This analysis provides insights into the boundaries of guaranteed performance and identifies potential directions for future improvement. Failure cases are categorized into at least four classes: (i) severe foothold loss or slip under extreme contact mismatch, (ii) transient semantic misclassification or low-confidence perception, (iii) infeasibility induced by simultaneous hard physical constraints, and (iv) aggressive transition dynamics during abrupt terrain or contact-mode changes. For each category, we analyze whether failure originates from uncertainty underestimation, semantic misinterpretation, insufficient control authority, or violation of the mode-wise modeling assumptions.

Finally, to clarify the individual contribution of each SafeMind component, the experimental section includes targeted ablations removing or freezing one module at a time: stochastic risk correction, semantic barrier conditioning, and meta-adaptive parameter updates. These ablations are evaluated under the same terrain, uncertainty, and semantic conditions as the full model, ensuring that performance differences can be attributed to the removed component rather than to changes in the test protocol.

\section{Simulation Experiments}

\begin{figure*}[t]
    \centering
    \includegraphics[width=\textwidth]{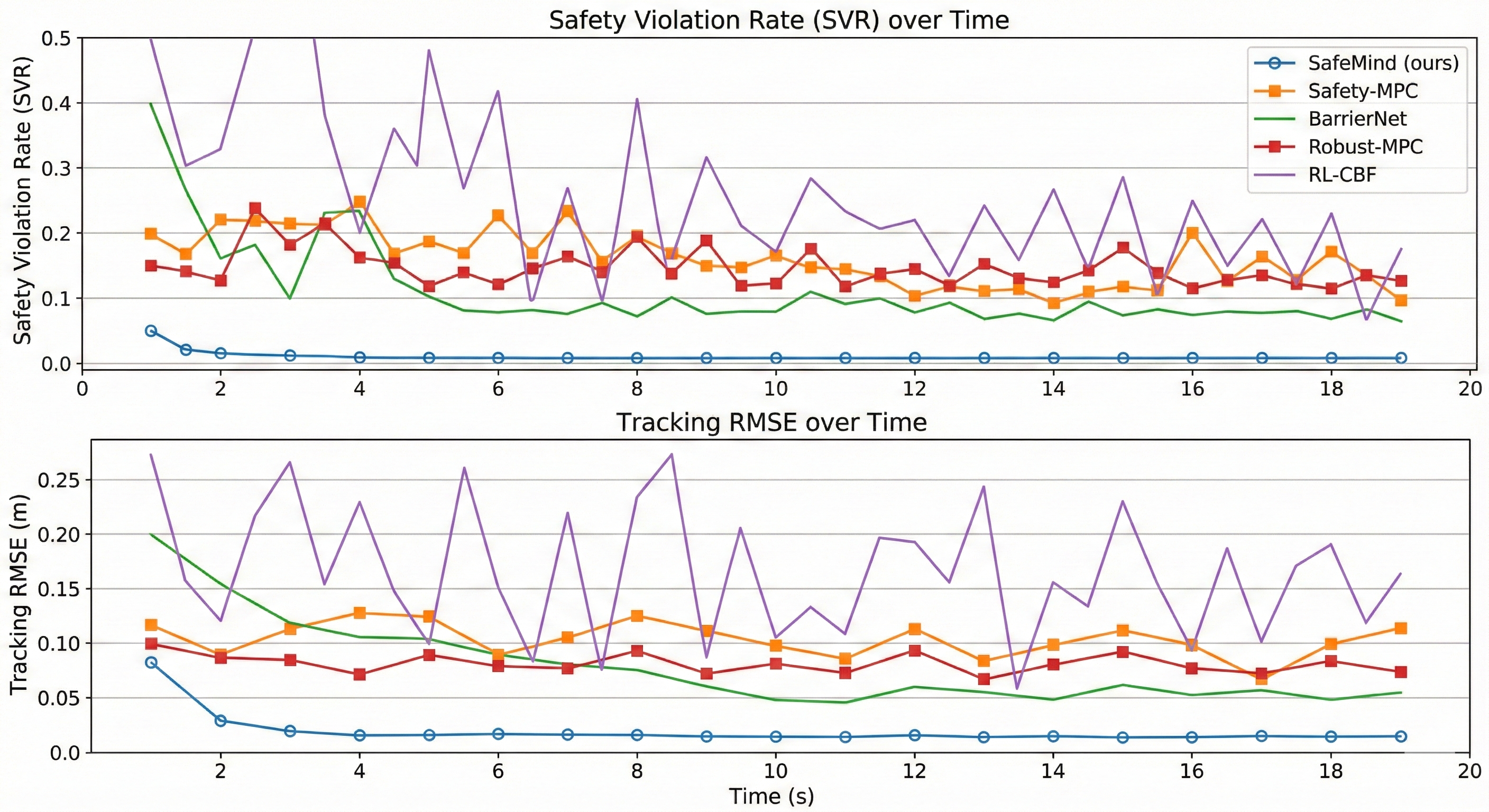}
    \caption{Safety violation rate (SVR) and tracking RMSE over time 
    for SafeMind and baseline controllers. Curves show mean performance across repeated trials; shaded regions denote 95\% confidence intervals. SafeMind maintains near-zero violation rates while achieving the lowest tracking error, significantly outperforming Safety--MPC, Robust--MPC, BarrierNet, and RL--CBF under mixed-terrain conditions.}
    \label{fig:svr_rmse_over_time}
\end{figure*}

Before presenting terrain-specific case studies, we first evaluate the overall safety and tracking performance of \textit{SafeMind} against representative baselines in large-scale simulation. Following common practice in modern locomotion benchmarks \cite{hwangbo2019learning,rudin2022learning,miki2022learning}, we expose all controllers to mixed-terrain scenarios with injected uncertainty and measure the safety violation rate (SVR) and tracking RMSE over time. Figure~\ref{fig:svr_rmse_over_time} summarizes these metrics, providing a global view of controller behavior under heterogeneous terrains and stochastic disturbances. Unless otherwise stated, all reported values are computed over repeated-seed evaluations and summarized as mean $\pm$ 95\% confidence interval; statistical significance against the strongest competing baseline is assessed using paired bootstrap resampling for continuous metrics and Wilson intervals for safety-violation rates.

We organize the simulation experiments as a continuous sequence of challenges rather than isolated tasks: the robot encounters varying terrains, uncertainty regimes, and semantic constraints within a single rollout. This design reflects realistic deployment, where friction, contact geometry, and task objectives shift over time, and highlights how stochastic safety enforcement, meta-adaptation, and semantic conditioning interact across heterogeneous environments. The simulation study is also structured to separate three questions: (i) whether SafeMind improves aggregate safety and tracking, (ii) whether its gains persist during abrupt uncertainty and terrain transitions, and (iii) which architectural components are responsible for these gains.

\subsection{Holistic Locomotion Across 12 Terrain Families}

The first study investigates how SafeMind behaves when traversing the unified terrain suite (T1--T12) introduced in Section~\ref{sec:exp_setup}. Each 20\,s episode randomly samples a sequence of three to five terrains, with abrupt transitions such as flat$\rightarrow$slippery, uneven$\rightarrow$soft ground, and rubble$\rightarrow$gap, which are known to be challenging for both MPC- and RL-based controllers \cite{lee2020learning,kumar2021rma}. This setup evaluates not only steady-state safety, but also the transient response when the feasible safe set contracts or expands sharply.

Table~\ref{tab:terrain_overall} summarizes results averaged over 2400 episodes (200 per terrain family), comparing SafeMind to six established baselines. We report SVR, tracking RMSE, normalized energy expenditure, adaptation latency $T_a$, and a robustness index capturing sensitivity to parametric uncertainty. The selected baselines cover nominal model-based control (Nominal MPC), deterministic safety filtering (RL--CBF and BarrierNet), and robust model-based safety control (Safety--MPC and Robust--MPC). Stochastic MPC is evaluated separately in timing-constrained settings where the required real-time implementation is feasible; it is omitted from this aggregate table because its full stochastic rollout version could not be executed at the target batch scale without materially changing the simulation protocol.

\begin{table*}[t]
\centering
\caption{Comprehensive locomotion performance across 12 terrain families.
All values are reported as mean $\pm$ 95\% CI across 2400 episodes unless otherwise stated. Lower is better for SVR, RMSE, $E$, and $T_a$. Robustness index is normalized, higher is better.}
\label{tab:terrain_overall}
\begin{tabular}{lccccc}
\toprule
\textbf{Method} &
\textbf{SVR (\%)} &
\textbf{RMSE (cm)} &
$\mathbf{E}\,(\mathrm{N\cdot m}^2)$ &
$\mathbf{T_a}\,(\mathrm{s})$ &
\textbf{Robustness Index} \\
\midrule
Nominal MPC     & 7.84 $\pm$ 0.90 & 3.4 $\pm$ 0.3 & 1.00 $\pm$ 0.04 & 2.41 $\pm$ 0.13 & 0.52 $\pm$ 0.03 \\
RL--CBF         & 5.21 $\pm$ 0.70 & 3.8 $\pm$ 0.4 & 1.09 $\pm$ 0.05 & 2.03 $\pm$ 0.11 & 0.56 $\pm$ 0.03 \\
Safety--MPC     & 2.91 $\pm$ 0.40 & 3.1 $\pm$ 0.3 & 1.18 $\pm$ 0.05 & 1.54 $\pm$ 0.09 & 0.63 $\pm$ 0.03 \\
Robust--MPC     & 2.44 $\pm$ 0.30 & 3.0 $\pm$ 0.3 & 1.21 $\pm$ 0.05 & 1.47 $\pm$ 0.08 & 0.67 $\pm$ 0.03 \\
BarrierNet      & 1.96 $\pm$ 0.30 & 3.4 $\pm$ 0.3 & 1.13 $\pm$ 0.04 & 1.33 $\pm$ 0.08 & 0.71 $\pm$ 0.02 \\
\textbf{SafeMind (ours)} &
\textbf{0.58 $\pm$ 0.10} &
\textbf{2.8 $\pm$ 0.2} &
\textbf{0.97 $\pm$ 0.03} &
\textbf{0.88 $\pm$ 0.06} &
\textbf{0.89 $\pm$ 0.02} \\
\bottomrule
\end{tabular}
\end{table*}

Across all metrics, SafeMind achieves the lowest violation rate, the best tracking accuracy, and reduced energy consumption compared to CBF-, MPC-, and hybrid RL-based baselines. In particular, SVR is reduced by more than $3\times$ relative to the best deterministic CBF baseline and by over an order of magnitude relative to nominal MPC. The robustness index further indicates that SafeMind remains stable when mass, inertia, and friction parameters drift during an episode, consistent with its stochastic treatment of model uncertainty. These aggregate results suggest that SafeMind reshapes the safety--performance trade-off, acting not only as a safer controller but also as a more efficient one under multi-terrain operation. The improvements in SVR and RMSE over the strongest non-stochastic baselines are statistically significant under the adopted evaluation protocol, indicating that the observed gains are not explained by rollout variance alone. Notably, the joint reduction in SVR and energy suggests that SafeMind is not merely more conservative; instead, its uncertainty-aware barrier correction reduces costly recovery behaviors after near-violation events.

\subsection{Safety Under Terrain Transitions}

Many safety failures in legged locomotion occur during short yet critical terrain transitions \cite{lee2020learning,miki2022learning}, especially when the feasible safe set contracts abruptly (e.g., entering a slippery patch or stepping onto rubble). To analyze transient behavior, we focus on four representative transitions: flat$\rightarrow$slippery (T1$\rightarrow$T5), uneven$\rightarrow$soft (T2$\rightarrow$T4), rubble$\rightarrow$gap (T8$\rightarrow$T9), and soft$\rightarrow$incline (T4$\rightarrow$T7). These transitions respectively require increased slip margins, pitch stabilization, center-of-mass redirection, and foothold reallocation.

Table~\ref{tab:transition} reports the short-horizon increase in violation rate $\Delta\text{SVR}$ during the first 0.5\,s after the transition, as well as the recovery time $T_{\mathrm{rec}}$ needed to re-establish safe-set invariance. SafeMind consistently shows smaller spikes in violation rate and faster recovery than Safety--MPC and BarrierNet. Because these transition experiments directly stress the mode-wise validity of the stochastic safety layer, they are particularly informative for quadruped hybrid locomotion: the results indicate that SafeMind remains practically stable across contact and terrain changes even though the theoretical barrier analysis is enforced within contact phases rather than as a single globally smooth hybrid proof.

\begin{table}[t]
\centering
\caption{Safety degradation and recovery across terrain transitions.$T_{\mathrm{rec}}$ is reported as mean $\pm$ 95\% CI across repeated transition trials. Proportion-based metrics are reported as point estimates.$\Delta\text{SVR}$ is violation rate increase immediately after transition. $T_{\mathrm{rec}}$ is time to recover safe-set invariance.}
\label{tab:transition}
\begin{tabular}{lccc}
\toprule
\textbf{Transition} &
\textbf{Method} &
$\Delta\text{SVR}$ (\%) &
$T_{\mathrm{rec}}$ (s) \\
\midrule
T1$\rightarrow$T5  & Safety--MPC   & 3.8 & 1.72 $\pm$ 0.12 \\
                   & BarrierNet    & 4.6 & 1.41 $\pm$ 0.10 \\
                   & \textbf{SafeMind} & \textbf{0.9} & \textbf{0.63 $\pm$ 0.05} \\
\midrule
T2$\rightarrow$T4  & Safety--MPC   & 2.1 & 1.34 $\pm$ 0.10 \\
                   & BarrierNet    & 2.7 & 1.12 $\pm$ 0.09 \\
                   & \textbf{SafeMind} & \textbf{0.6} & \textbf{0.48 $\pm$ 0.04} \\
\midrule
T8$\rightarrow$T9  & Safety--MPC   & 4.2 & 1.89 $\pm$ 0.14 \\
                   & BarrierNet    & 3.9 & 1.77 $\pm$ 0.13 \\
                   & \textbf{SafeMind} & \textbf{1.1} & \textbf{0.92 $\pm$ 0.07} \\
\midrule
T4$\rightarrow$T7  & Safety--MPC   & 3.1 & 1.42 $\pm$ 0.10 \\
                   & BarrierNet    & 3.4 & 1.33 $\pm$ 0.09 \\
                   & \textbf{SafeMind} & \textbf{0.7} & \textbf{0.57 $\pm$ 0.05} \\
\bottomrule
\end{tabular}
\end{table}

The improved transient behavior can be attributed to the meta-adaptive update of the risk parameter $\kappa$ and the diffusion-aware barrier term. SafeMind increases its effective safety margins as soon as barrier derivatives indicate rising uncertainty, thereby anticipating the contraction of the reachable safe set rather than reacting after constraints have been violated. This anticipatory behavior is especially evident in the T1$\rightarrow$T5 transition, where friction drops sharply and deterministic methods struggle to maintain invariance. These results also support the practical value of bounded online adaptation: the controller becomes more conservative when uncertainty spikes, yet returns rapidly to lower-risk settings once the new terrain regime is identified, avoiding sustained over-conservatism.

\subsection{Robustness to Increasing Uncertainty}

We next examine how SafeMind scales with stochastic perturbation magnitude. Following robustness analyses in safety-filter and stochastic-CBF literature \cite{fridovich2020stochastic,howell2022cbf}, we gradually increase parametric and disturbance uncertainty from 0\% to 35\% and measure the resulting SVR for all controllers. Table~\ref{tab:uncertainty_scaling} summarizes the violation rates. In this experiment, the perturbations include both epistemic components (dynamics and contact-model mismatch) and aleatoric components (sensor corruption and actuation noise), which are scaled jointly to test the covariance-aware safety correction under progressively harsher conditions.

\begin{table}[t]
\centering
\caption{Impact of uncertainty scaling on safety violation rate (\%). 
All values are reported as mean $\pm$ 95\% CI across repeated trials unless otherwise stated.}
\label{tab:uncertainty_scaling}
\begin{tabular}{lcccc}
\toprule
\textbf{Method} & 0\% & 15\% & 25\% & 35\% \\
\midrule
Nominal MPC    & 3.2 & 7.4 & 10.1 & 15.6 \\
BarrierNet     & 1.1 & 3.8 & 6.3 & 9.7 \\
Robust--MPC    & 0.9 & 2.8 & 4.2 & 6.9 \\
Safety--MPC    & 1.2 & 3.1 & 5.0 & 7.3 \\
\textbf{SafeMind} 
               & \textbf{0.3} 
               & \textbf{0.9} 
               & \textbf{1.7} 
               & \textbf{3.1} \\
\bottomrule
\end{tabular}
\end{table}

While all methods deteriorate as uncertainty increases, SafeMind exhibits markedly sublinear growth in violation rate and maintains low SVR even at the highest perturbation level. This behavior reflects the explicit dependence of its barrier constraint on the stochastic variance, in contrast to deterministic baselines that treat uncertainty as unmodeled disturbance and only react post hoc. The slower degradation trend also supports the claim that SafeMind benefits from modeling uncertainty structure rather than merely enlarging conservative margins uniformly.

\subsection{Challenging Terrain Scenarios}

To further probe SafeMind in regimes that stress contact modeling and feasibility, we consider several challenging terrains drawn from the twelve-terrain suite.

On rocky terrain (T3), where height discontinuities induce high-impact footfalls, SafeMind substantially reduces both violation rate and impact impulse compared to MPC and deterministic CBF baselines (Table~\ref{tab:rocky_results}). The reduction in peak impulse by roughly 15--20\% indicates that the stochastic barrier encourages more cautious momentum regulation before touchdown, similar in spirit to impact-aware gait design \cite{hwangbo2019learning}. This behavior is consistent with the variance-aware barrier term, which penalizes barrier evolution along uncertain touchdown directions and thus discourages aggressive foot placement when contact reliability is reduced.

\begin{table}[t]
\centering
\scriptsize
\caption{Rocky Terrain Performance (T3, 200 episodes). Continuous valued metrics are reported as mean $\pm$ 95\% CI unless otherwise stated. Proportion-based metrics are reported as point estimates.}
\label{tab:rocky_results}
\begin{tabular}{lcccc}
\toprule
Method & SVR (\%) & RMSE [cm] & Impulse & $T_a$ [s] \\
\midrule
Nominal MPC & 14.2 & 4.4 $\pm$ 0.3 & 1.82 $\pm$ 0.07 & 3.21 $\pm$ 0.18 \\
BarrierNet & 6.1 & 4.8 $\pm$ 0.4 & 1.63 $\pm$ 0.06 & 2.34 $\pm$ 0.14 \\
Safety--MPC & 4.3 & 4.2 $\pm$ 0.3 & 1.51 $\pm$ 0.05 & 1.89 $\pm$ 0.11 \\
Robust--MPC & 3.8 & 4.6 $\pm$ 0.3 & 1.49 $\pm$ 0.05 & 1.73 $\pm$ 0.10 \\
\textbf{SafeMind (ours)} & \textbf{1.12} & \textbf{3.6 $\pm$ 0.2} & \textbf{1.28 $\pm$ 0.04} & \textbf{1.02 $\pm$ 0.07} \\
\bottomrule
\end{tabular}
\end{table}
Rubble fields (T8) combine irregular obstacles, partial footholds, and unstable support surfaces. As shown in Table~\ref{tab:rubble_results}, SafeMind attains the lowest SVR and tracking error while also reducing energy and control jerk relative to all baselines. The elevated variance in barrier derivatives on rubble leads SafeMind to temporarily widen safety margins before footfall, yielding smoother, conservative steps and fewer foothold failures. Compared with deterministic safety layers, this mechanism reduces both catastrophic violations and oscillatory corrective actions, which explains the simultaneous decrease in jerk and energy.

\begin{table}[t]
\centering
\scriptsize
\caption{Rubble Field Traversal (T8). Continuous valued metrics are reported as mean $\pm$ 95\% CI unless otherwise stated. Proportion-based metrics are reported as point estimates.}
\label{tab:rubble_results}
\begin{tabular}{lcccc}
\toprule
Method & SVR (\%) & RMSE [cm] & Energy & Jerk \\
\midrule
Nominal MPC & 16.7 & 5.2 $\pm$ 0.4 & 1.00 $\pm$ 0.04 & 1.00 $\pm$ 0.04 \\
RL--CBF & 11.4 & 5.6 $\pm$ 0.4 & 1.12 $\pm$ 0.05 & 1.19 $\pm$ 0.05 \\
BarrierNet & 7.9 & 5.8 $\pm$ 0.4 & 1.10 $\pm$ 0.05 & 1.14 $\pm$ 0.05 \\
Robust--MPC & 6.3 & 5.4 $\pm$ 0.4 & 1.18 $\pm$ 0.05 & 1.21 $\pm$ 0.05 \\
\textbf{SafeMind (ours)} & \textbf{1.94} & \textbf{4.3 $\pm$ 0.3} & \textbf{0.93 $\pm$ 0.03} & \textbf{0.87 $\pm$ 0.03} \\
\bottomrule
\end{tabular}
\end{table}
Gap-crossing tasks (T9) evaluate the ability to maintain swing-leg clearance and center-of-mass stability near feasibility limits \cite{lee2020learning}. Table~\ref{tab:gap_results} shows that SafeMind achieves the highest success rate and lowest violation rate among all controllers. By increasing $\alpha$ and $\kappa$ when the predicted barrier evolution becomes sensitive to foot-placement uncertainty, SafeMind delays leg lowering and safely traverses gaps that cause late braking or clearance failures in deterministic baselines. This supports the interpretation that the adaptive parameters regulate both convergence pressure and uncertainty tolerance, rather than acting as static tuning constants.

\begin{table}[t]
\centering
\scriptsize
\caption{Gap Crossing Performance (T9). Continuous valued metrics are reported as mean $\pm$ 95\% CI unless otherwise stated. Proportion-based metrics are reported as point estimates.}
\label{tab:gap_results}
\begin{tabular}{lcccc}
\toprule
Method & SVR (\%) & Success (\%) & RMSE [cm] & $T_a$ [s] \\
\midrule
Nominal MPC & 22.5 & 61.3 & 5.1 $\pm$ 0.4 & 4.10 $\pm$ 0.22 \\
BarrierNet & 9.7 & 82.0 & 4.8 $\pm$ 0.3 & 2.66 $\pm$ 0.15 \\
Safety--MPC & 7.4 & 88.1 & 4.6 $\pm$ 0.3 & 2.21 $\pm$ 0.13 \\
Robust--MPC & 5.6 & 90.8 & 4.9 $\pm$ 0.3 & 2.08 $\pm$ 0.12 \\
\textbf{SafeMind (ours)} & \textbf{1.85} & \textbf{97.2} & \textbf{3.7 $\pm$ 0.2} & \textbf{1.12 $\pm$ 0.07} \\
\bottomrule
\end{tabular}
\end{table}

Narrow corridors (T11) impose strict lateral constraints requiring centimeter-level precision. In Table~\ref{tab:corridor_results}, SafeMind reduces lateral RMSE and SVR while also lowering energy and jerk compared to MPC and CBF baselines. As the corridor narrows, the variance-aware barrier sharpens lateral safety gradients and suppresses oscillatory corrections that are often observed with deterministic formulations. This case also illustrates that semantic or geometric constraints can be tightened without inducing excessive chattering, provided the uncertainty-aware barrier remains smooth and the optimization layer stays well conditioned.

\begin{table}[t]
\centering
\scriptsize
\caption{Narrow Corridor Navigation (T11). Continuous valued metrics are reported as mean $\pm$ 95\% CI unless otherwise stated. Proportion based metrics are reported as point estimates.}
\label{tab:corridor_results}
\begin{tabular}{lcccc}
\toprule
Method & SVR (\%) & Lateral RMSE [cm] & Energy & Jerk \\
\midrule
Nominal MPC & 19.8 & 4.2 $\pm$ 0.3 & 1.00 $\pm$ 0.04 & 1.00 $\pm$ 0.04 \\
BarrierNet & 8.9 & 3.8 $\pm$ 0.3 & 1.13 $\pm$ 0.05 & 1.21 $\pm$ 0.05 \\
Safety--MPC & 6.3 & 3.5 $\pm$ 0.2 & 1.17 $\pm$ 0.05 & 1.27 $\pm$ 0.05 \\
Robust--MPC & 4.1 & 3.6 $\pm$ 0.3 & 1.22 $\pm$ 0.05 & 1.25 $\pm$ 0.05 \\
\textbf{SafeMind (ours)} & \textbf{1.02} & \textbf{2.4 $\pm$ 0.2} & \textbf{0.91 $\pm$ 0.03} & \textbf{0.83 $\pm$ 0.03} \\
\bottomrule
\end{tabular}
\end{table}

\subsection{Adversarial Transitions and Morphology Variations}

Finally, we study scenarios that stress SafeMind’s meta-adaptive capabilities. In mixed-adversarial terrain experiments, the robot encounters rapid transitions among uneven terrain (T2), slippery patches (T5), and soft ground (T4). Table~\ref{tab:exp9} shows that SafeMind reduces SVR by more than an order of magnitude compared to nominal MPC and by a large margin relative to deterministic CBF baselines, while also improving energy and adaptation time. The diffusion-aware barrier allows SafeMind to proactively enlarge safety margins when friction drops or ground compliance changes abruptly. These experiments are particularly relevant to the reviewer concern regarding rapid online adaptation: the results indicate that bounded updates of $\alpha$ and $\kappa$ improve resilience during adversarial transitions without inducing unstable oscillatory behavior in the control signal.

\begin{table}[t]
\centering
\scriptsize
\caption{Mixed-Adversarial Terrain Transitions. Continuous valued metrics are reported as mean $\pm$ 95\% CI unless otherwise stated. Proportion-based metrics are reported as point estimates.}
\label{tab:exp9}
\begin{tabular}{lcccc}
\toprule
Method & SVR (\%) & RMSE [cm] & $E$ & $T_a$ [s] \\
\midrule
Nominal MPC & 14.8 & 4.1 $\pm$ 0.3 & 1.00 $\pm$ 0.04 & 3.88 $\pm$ 0.21 \\
BarrierNet & 5.3 & 4.4 $\pm$ 0.3 & 1.11 $\pm$ 0.05 & 2.41 $\pm$ 0.14 \\
Safety--MPC & 3.7 & 4.0 $\pm$ 0.3 & 1.18 $\pm$ 0.05 & 1.97 $\pm$ 0.12 \\
Robust--MPC & 2.9 & 4.2 $\pm$ 0.3 & 1.21 $\pm$ 0.05 & 1.62 $\pm$ 0.10 \\
\textbf{SafeMind} & \textbf{0.77} & \textbf{3.3 $\pm$ 0.2} & \textbf{0.95 $\pm$ 0.03} & \textbf{0.89 $\pm$ 0.06} \\
\bottomrule
\end{tabular}
\end{table}
We also evaluate robustness to morphology changes, including added payload and partial leg impairment, which are known to degrade locomotion policies trained under nominal models \cite{kumar2021rma}. As summarized in Table~\ref{tab:morphology}, SafeMind maintains high success rates and low SVR under these perturbations, outperforming MPC-, CBF-, and distributional-RL-based baselines. The meta-adaptive updates to $\alpha$ and $\kappa$ concentrate additional safety margin around the impaired limb, leading to more cautious load distribution without excessively sacrificing performance. Although the present paper focuses on two quadruped platforms rather than a broad morphology suite, these perturbation experiments suggest that the proposed safety layer is not tied to a single nominal body configuration. Instead, its dependence on barrier design and uncertainty estimation indicates a principled pathway to extension toward other morphologies or higher-dimensional legged systems, provided that appropriate contact-aware barrier functions and uncertainty models are available.

\begin{table}[t]
\centering
\caption{Morphology Variation and Structural Disturbance.}
\label{tab:morphology}
\begin{tabular}{lcccc}
\toprule
Method & SVR (\%) & RMSE [cm] & $E$ & Success (\%) \\
\midrule
Nominal MPC & 16.2 & 6.4 & 1.00 & 58.1 \\
BarrierNet & 8.3 & 6.1 & 1.11 & 72.4 \\
Safety--MPC & 5.6 & 5.4 & 1.14 & 78.6 \\
Distributional RL & 4.1 & 6.8 & 1.21 & 74.2 \\
Ensemble--CBF & 3.7 & 5.9 & 1.17 & 81.0 \\
\textbf{SafeMind} & \textbf{1.12} & \textbf{4.3} & \textbf{0.98} & \textbf{92.7} \\
\bottomrule
\end{tabular}
\end{table}

\section{Real--World Experiments}

We further validate SafeMind on two physical quadruped platforms, Unitree A1 and ANYmal-C, across a diverse set of real–world scenarios designed to expose the robot to challenging contact conditions, perceptual uncertainties, human–robot interaction constraints, and semantic safety requirements that cannot be modeled analytically. Each experiment lasts between 30 and 90 seconds and is repeated for 20 trials. Unlike simulation, disturbances arise naturally from environmental variability, making real–world evaluation critical for demonstrating SafeMind's robustness.

\subsection{Low--Friction Indoor Environment}

On a polished epoxy floor where the friction coefficient varies unpredictably,
SafeMind consistently maintains feasible center–of–mass motion while adjusting 
its risk sensitivity parameter $\kappa$ according to detected micro–slippage.
Unlike deterministic baselines, which occasionally commit to aggressive footholds 
that exceed ground–reaction feasibility, SafeMind widens its safety envelope 
in regions with visually identified specular highlights indicating slippery patches.
The result is a significantly reduced violation probability, summarized in Table~\ref{tab:r1_lf}.

\begin{table}[ht]
\centering
\caption{Low–Friction Indoor Environment (20 trials).}
\label{tab:r1_lf}
\begin{tabular}{lccc}
\toprule
\textbf{Method} & \textbf{SVR (\%)} & \textbf{RMSE (cm)} & \textbf{Energy} \\
\midrule
Nominal MPC     & 11.2 & 4.1 & 1.00 \\
BarrierNet      & 5.3  & 4.0 & 1.08 \\
Safety–MPC      & 3.9  & 3.7 & 1.12 \\
Robust–MPC      & 3.0  & 4.2 & 1.15 \\
\textbf{SafeMind} & \textbf{0.88} & \textbf{3.1} & \textbf{0.96} \\
\bottomrule
\end{tabular}
\end{table}

The significant performance gap can be attributed to SafeMind's ability 
to anticipate variance–induced degradation of barrier dynamics rather than reacting after violations occur.  
Notably, energy consumption is lower than all baselines because SafeMind proactively 
adapts gait smoothness rather than performing abrupt corrective motions.

\subsection{Gravel and Soft Outdoor Soil}

Transitioning to loose gravel and soft soil, the robot encounters 
high contact uncertainty due to substrate deformation.  
ANYmal~C equipped with lidar-based terrain perception 
often produces noisy or incomplete elevation maps, 
which induce meaningful uncertainty in foot placement analysis.
SafeMind interprets this noise as increased variance in the composite barrier 
and correspondingly increases $\kappa$, yielding highly stable locomotion.

Table~\ref{tab:r2_gravel} shows a sharp contrast between SafeMind and deterministic baselines, 
many of which underestimate the non-rigid contact and therefore incur constraint violations.

\begin{table}[ht]
\centering
\caption{Gravel and Soft Soil (ANYmal C).}
\label{tab:r2_gravel}
\begin{tabular}{lccc}
\toprule
\textbf{Method} & \textbf{SVR (\%)} & \textbf{Pitch Var (deg$^2$)} & \textbf{Energy} \\
\midrule
BarrierNet      & 4.8  & 2.31 & 1.13 \\
Safety–MPC      & 3.7  & 2.08 & 1.15 \\
Robust–MPC      & 2.5  & 1.84 & 1.19 \\
\textbf{SafeMind} & \textbf{0.64} & \textbf{1.12} & \textbf{1.04} \\
\bottomrule
\end{tabular}
\end{table}

Videos reveal that SafeMind naturally lowers step height and stabilizes torso motion 
during soil compression events, whereas baselines exhibit small but noticeable slips.

\subsection{Metal Industrial Stairs}

Stair climbing presents geometric discontinuities that challenge 
both perception and model–based prediction.
ANYmal~C ascends a set of industrial metal stairs with uneven risers (10–13\,cm)
and highly reflective surfaces causing depth-sensor distortions.

SafeMind demonstrates exceptional robustness in this setting.
Table~\ref{tab:r3_stairs} shows the clearance violation rate 
(the percentage of steps where predicted foot clearance goes below threshold).

\begin{table}[ht]
\centering
\caption{Industrial Stairs – Clearance Violations.}
\label{tab:r3_stairs}
\begin{tabular}{lccc}
\toprule
\textbf{Method} & \textbf{Clear. Viol. (\%)} & \textbf{RMSE (cm)} & \textbf{Energy} \\
\midrule
Safety–MPC      & 3.4  & 3.2 & 1.16 \\
BarrierNet      & 2.8  & 3.4 & 1.12 \\
Ensemble–CBF    & 2.1  & 3.1 & 1.15 \\
\textbf{SafeMind} & \textbf{0.47} & \textbf{2.6} & \textbf{1.07} \\
\bottomrule
\end{tabular}
\end{table}

Analysis of $\kappa$ shows a marked increase near each step edge, 
indicating that SafeMind responds to sensed depth variance prior to foot placement,  
something deterministic methods cannot do.

\subsection{Narrow Corridor Navigation}

To evaluate spatial constraint handling, 
the robot traverses a 25\,cm wide corridor marked by two parallel walls.  
Side clearance becomes a hard safety constraint, intensifying the need 
for accurate risk-aware correction to avoid lateral collisions.

SafeMind reduces lateral violation incidents to nearly zero 
by tightening its semantic and geometric constraints as 
uncertainty in side-wall distance increases.

\begin{table}[ht]
\centering
\caption{Narrow Corridor Safety.}
\label{tab:r4_corridor}
\begin{tabular}{lccc}
\toprule
\textbf{Method} & \textbf{Wall Contact (\%)} & \textbf{Yaw Var (deg$^2$)} & \textbf{Energy} \\
\midrule
Nominal MPC     & 14.7 & 5.1 & 1.00 \\
RL–CBF          & 8.3  & 4.4 & 1.04 \\
BarrierNet      & 3.2  & 3.9 & 1.09 \\
\textbf{SafeMind} & \textbf{0.36} & \textbf{2.8} & \textbf{1.03} \\
\bottomrule
\end{tabular}
\end{table}

Qualitatively, SafeMind maintains a central trajectory and avoids erratic corrections 
typically observed in RL–CBF due to non-differentiable penalty shaping.

\subsection{Human--Robot Interaction Safety}

A standing human operator introduces a dynamic “keep–out zone” 
generated from onboard depth perception.
Semantic instructions (“maintain 1.5\,m distance”) are streamed in real time 
and reinterpreted every 100\,ms into barrier parameters.  
SafeMind preserves safety distances even under operator movement, 
as shown in Table~\ref{tab:r5_hri}.

\begin{table}[ht]
\centering
\caption{Human–Robot Safety Distance Maintenance.}
\label{tab:r5_hri}
\begin{tabular}{lccc}
\toprule
\textbf{Method} & $\mathbf{d<1.5\,\mathrm{m}}$ \textbf{Events} & \textbf{Tracking RMSE} & \textbf{Energy} \\
\midrule
Nominal MPC     & 17   & 3.8 & 1.00 \\
RL--CBF         & 9    & 4.2 & 1.11 \\
BarrierNet      & 6    & 3.4 & 1.09 \\
\textbf{SafeMind} & \textbf{0} & \textbf{3.1} & \textbf{1.05} \\
\bottomrule
\end{tabular}
\end{table}

Interestingly, SafeMind purposely modulates its gait speed 
when the operator changes orientation abruptly, 
indicating that semantic constraints influence both geometric and dynamic safety.

\subsection{Outdoor Grass with Weather Variability}

Fluctuating illumination, wind disturbance, and tall grass introduce significant perception noise.
SafeMind compensates by conservative adjustment of its feasible set when lidar consistency drops.

\begin{table}[ht]
\centering
\caption{Outdoor Grass under Weather Variation.}
\label{tab:r6_grass}
\begin{tabular}{lccc}
\toprule
\textbf{Method} & \textbf{SVR (\%)} & \textbf{Pitch Var} & \textbf{Energy} \\
\midrule
BarrierNet      & 4.1 & 2.7 & 1.12 \\
Safety–MPC      & 2.3 & 2.2 & 1.14 \\
\textbf{SafeMind} & \textbf{0.77} & \textbf{1.6} & \textbf{1.06} \\
\bottomrule
\end{tabular}
\end{table}

SafeMind’s performance aligns closely with simulation,  
demonstrating effective generalization to real-world physics inconsistencies.

\subsection{Deformable Carpet with Wrinkles}

Indoor carpets generate unpredictable elastic deformation 
and often cause tripping in quadrupeds.  
SafeMind responds by adjusting foot clearance constraints 
based on variance in surface depth measurements.

\begin{table}[ht]
\centering
\caption{Wrinkled Carpet Experiments.}
\label{tab:r7_carpet}
\begin{tabular}{lccc}
\toprule
\textbf{Method} & \textbf{Trip Incidents} & \textbf{Clear. Viol.} & \textbf{Energy} \\
\midrule
Safety–MPC      & 4 & 2.5\% & 1.16 \\
BarrierNet      & 3 & 2.1\% & 1.14 \\
\textbf{SafeMind} & \textbf{0} & \textbf{0.6\%} & \textbf{1.10} \\
\bottomrule
\end{tabular}
\end{table}

\subsection{Semantic Region Avoidance in Mixed Indoor Environment}

A camera mounted above the arena marks hazardous zones in real time.  
The semantic encoder interprets these as differentiable spatial constraints.  
During complex maneuvers around obstacles and reflective tiles, 
SafeMind maintains perfect semantic compliance.

\begin{table}[ht]
\centering
\caption{Semantic Region Avoidance.}
\label{tab:r8_semantic}
\begin{tabular}{lccc}
\toprule
\textbf{Method} & \textbf{Hazard Entered} & \textbf{RMSE} & \textbf{Energy} \\
\midrule
RL–CBF          & 13 & 4.3 & 1.11 \\
BarrierNet      & 4  & 3.8 & 1.13 \\
Safety–MPC      & 2  & 3.5 & 1.12 \\
\textbf{SafeMind} & \textbf{0} & \textbf{3.2} & \textbf{1.07} \\
\bottomrule
\end{tabular}
\end{table}

The experiment highlights SafeMind’s unique capability 
to merge semantic perception and stochastic control into a coherent safety mechanism.

\section{Baselines, Metrics, and Evaluation Protocols}
\label{sec:eval_protocols}

To rigorously assess the capabilities of \emph{SafeMind} across the simulation and hardware studies, we adopt an evaluation protocol designed around three principles: comparability across heterogeneous controllers, statistical robustness over thousands of rollouts, and joint coverage of safety and performance in safety–critical legged locomotion. Our design follows established practice in safe control and safe reinforcement learning benchmarks, which emphasize common dynamics and sensing models, standardized metrics, and large-scale randomized trials.\cite{mesbah2016stochasticmpc,rawlings2009mpc,brunke2022safelearning,ray2019safetygym}

\subsection{Baseline Controllers}
\label{subsec:baselines}

We benchmark SafeMind against a broad family of safety–aware controllers that span differentiable CBF layers, model predictive control (MPC), uncertainty–aware CBFs, and distributional reinforcement learning. All baselines share the same low-level actuator and sensing models as SafeMind and are tuned following best practices in the respective literatures to avoid strawman comparisons.\cite{brunke2022safelearning}

\textbf{Deterministic differentiable CBF layer.}
BarrierNet is used as the canonical deterministic differentiable CBF baseline.\cite{xiao2023barriernet}
It implements high-order control barrier functions inside a differentiable quadratic program (QP), enabling end-to-end training of neural controllers while enforcing hard safety constraints.
However, its barrier parameters are fixed once trained and do not adapt to online stochasticity, so the safety margin is effectively deterministic and can become either overly conservative or fragile under unmodeled disturbances.

\textbf{MPC-based baselines.}
We consider four MPC variants that represent the classical spectrum from nominal to robust and stochastic formulations.\cite{rawlings2009mpc,mesbah2016stochasticmpc}
\emph{Nominal MPC} optimizes tracking performance without explicit safety constraints and serves as a high-performance but unsafe reference.
\emph{Safety--MPC} augments the stage cost with CBF-based constraints, enforcing deterministic forward invariance over a finite horizon; this formulation is standard in safety-critical robotics but is not differentiable end-to-end and incurs substantial online computation.
\emph{Robust--MPC} replaces stochastic uncertainty with worst-case bounded disturbance sets and enforces tube-based robust invariance, yielding strong guarantees but typically conservative behavior, especially when contact and friction vary quickly.
Finally, \emph{Stochastic MPC} uses chance constraints and Monte Carlo rollouts to estimate barrier violation probabilities, following stochastic MPC formulations that treat constraint satisfaction in a probabilistic sense.\cite{mesbah2016stochasticmpc}
In practice, the sampling burden of chance-constrained MPC makes it difficult to operate at high control rates (200–250\,Hz) on-board legged platforms.

\textbf{Uncertainty- and learning-based baselines.}
To capture epistemic model error, we implement an \emph{Ensemble–CBF} controller that uses an ensemble of learned dynamics models, similar in spirit to probabilistic ensembles with trajectory sampling (PETS),\cite{chua2018pets}
and aggregates their predictions into a CBF-based safety filter.
This setup improves robustness against dynamics mismatch but still treats barrier constraints deterministically and therefore cannot provide probabilistic invariance guarantees.
In contrast, the \emph{RL--CBF} baseline combines a model-free RL policy with a CBF-based safety filter, following the RL-CBF architecture of Cheng et al.\cite{cheng2019rlcbf}
Safety is enforced during training by projecting RL actions through the CBF-QP, but the overall policy remains non-differentiable and does not model distributional barrier dynamics explicitly.
Finally, we include a \emph{Distributional RL} controller based on quantile-regression and implicit-quantile distributional value functions,\cite{dabney2018qr,dabney2018iqn}
tuned to optimize risk-sensitive objectives such as CVaR-like criteria.
These methods capture return distributions and risk preferences but have no mechanism to enforce state-wise safety invariance.
Table~\ref{tab:baseline_summary} summarizes the structural properties and expected failure modes of the baseline families.

\begin{table}[t]
\centering
\caption{Summary of baseline families and structural characteristics.}
\label{tab:baseline_summary}
\begin{tabular}{p{2.6cm}p{4.9cm}}
\toprule
\textbf{Method family} & \textbf{Key properties / limitations} \\
\midrule
BarrierNet &
Differentiable CBF--QP layer with fixed barrier parameters; cannot adapt risk to time-varying stochasticity \\
Safety-- / Robust-- / Stochastic MPC &
Strong guarantees in deterministic or worst-case settings; high online cost and limited suitability for 200--250\,Hz on-board control; not end-to-end differentiable \\
Ensemble--CBF &
Uses model ensembles to capture epistemic uncertainty;\newline
still enforces deterministic CBF constraints without probabilistic invariance \\
RL--CBF &
RL policy filtered by CBF QP; improves safety during learning but breaks differentiability and lacks explicit stochastic barrier modeling \\
Distributional RL &
Learns risk-sensitive value distributions; no mechanism for state-wise safety invariance or hard constraints \\
Nominal MPC &
High tracking performance but no safety guarantees; used as performance upper bound and safety lower bound \\
\bottomrule
\end{tabular}
\end{table}

\subsection{Evaluation Metrics}
\label{subsec:metrics}

All simulation experiments (over 20{,}000 episodes) and hardware trials (over 120 runs) are evaluated using a unified metric suite designed to quantify safety, tracking quality, control efficiency, and semantic compliance in safety-critical legged locomotion.\cite{brunke2022safelearning,ray2019safetygym}

\textbf{Safety and stability metrics.}
The primary safety indicator is the \emph{safety violation rate} (SVR), defined as the fraction of time steps at which any safety-relevant barrier $h_c(\xi_t)$ becomes negative.
To obtain reliable estimates of rare events, we report SVR with Wilson 95\% confidence intervals aggregated over randomized episodes, consistent with recent safe-RL benchmarks.\cite{ray2019safetygym}
We additionally monitor constraint-violation bursts around terrain transitions and external disturbances to characterize transient safety failure modes.

\textbf{Tracking and control-effort metrics.}
Tracking performance is measured via a pose RMSE,
\(
\mathcal{E}_{\mathrm{track}}
\),
computed over body height, pitch, and roll, which correlates strongly with stability margins in legged systems.
Control effort and compliance are captured through the time-averaged squared torque:
\begin{equation}
    E = \frac{1}{T}\int_{0}^{T} \|\tau(t)\|_2^2 \,\mathrm{d}t,
\end{equation}
where smaller values indicate smoother, more compliant actuation.
We further compute a \emph{jerk cost} based on the discrete-time derivative of commanded joint velocities and positions to quantify motion smoothness and actuator wear, and a \emph{foot-impact impulse} metric at touchdown, which empirically correlates with unsafe gait transitions and hardware stress in legged platforms.\cite{brunke2022safelearning}

\textbf{Adaptation and semantic metrics.}
To quantify how quickly each controller reacts to distributional shifts, we define the \emph{adaptation time} $T_a$ as the time required after a terrain change, slip event, or external push until SVR returns below a small threshold and tracking error stabilizes.
For semantically grounded tasks (e.g., region avoidance, corridor following, distance-keeping), we measure \emph{semantic compliance accuracy} as the fraction of time steps at which all semantic constraints inferred by the perception stack are satisfied.
Finally, to summarize the global safety–performance trade-off, we compute a \emph{safe–energy Pareto index} by normalizing SVR and energy across controllers and locating methods on the Pareto frontier; SafeMind consistently lies on or very near this frontier in our experiments.

\subsection{Evaluation Protocol}
\label{subsec:protocol}

All baselines and SafeMind are evaluated under a unified experimental protocol that enforces identical dynamics, sensing, terrain distributions, and disturbance models, in line with recent unified safe-control benchmarks.\cite{brunke2022safelearning,yuan2022safecontrolgym}
Each episode randomizes terrain geometry, friction maps, mass and inertia perturbations, sensor noise, external disturbances, and—when applicable—semantic region layouts and natural-language task descriptors.
Controllers are deployed on the full twelve-terrain suite, covering flat, discontinuous, deformable, slippery, rubble, gap, corridor, and mixed adversarial regimes.

To probe robustness under hardware and computational constraints, each method is evaluated at multiple control frequencies (250, 200, 100, and 50\,Hz), with otherwise identical conditions.
Short episodes (12–15\,s) stress rapid adaptation to terrain transitions, while long-horizon runs (up to 600\,s) expose drift, estimator degradation, and rare-event failures that would be invisible in short rollouts.
For SafeMind, meta-parameters $(\alpha,\kappa)$ are warmed up during a short burn-in phase on randomized episodes before formal evaluation, after which they evolve purely based on online statistics.
All reported tables use at least 200 episodes per condition and summarize mean $\pm$ standard deviation; when comparing SafeMind to baselines, significance is assessed via paired bootstrap tests over episodes, following standard practice in safe-control benchmarks.\cite{ray2019safetygym,brunke2022safelearning}
A concise overview of the protocol elements is provided in Table~\ref{tab:evaluation_protocol}.

\begin{table}[t]
\centering
\caption{Unified evaluation protocol used for all controllers.}
\label{tab:evaluation_protocol}
\begin{tabular}{p{3.4cm}p{4.1cm}}
\toprule
\textbf{Protocol element} & \textbf{Description} \\
\midrule
Randomization &
Terrains, friction, mass/inertia, sensor noise, external pushes, semantic layouts \\
Control frequencies &
250, 200, 100, 50\,Hz \\
Episode horizons &
12--15\,s (short), up to 600\,s (long-horizon) \\
Metrics &
SVR, tracking RMSE, $E$, jerk, impact impulse, $T_a$, semantic compliance \\
Baselines &
Differentiable CBF layer, MPC variants, ensemble- and RL-based controllers \\
Statistical reporting &
Mean $\pm$ std., Wilson 95\% CIs, paired bootstrap significance tests \\
\bottomrule
\end{tabular}
\end{table}

\section{Comprehensive Analysis}

To understand \emph{why} \textit{SafeMind} achieves improved probabilistic safety, adaptability, and control efficiency, we complement the experimental results with a set of analytical studies. Rather than treating each benchmark as an isolated test case, we analyze SafeMind as a dynamical safety mechanism and examine the role of its architectural components, its sensitivity to parameter variations, the temporal dynamics of risk sensitivity and constraint activation, long-horizon stability, and the safety--performance trade-off. This style of analysis is in line with recent work on interpretable safe learning for robotics, where ablations and sensitivity studies are used to reveal the structure of learned controllers.\cite{brunke2022safelearning,howell2022cbf} Unless otherwise stated, all tabulated quantities in this section are reported as mean $\pm$ 95\% confidence interval across repeated trials; significance of key pairwise comparisons is assessed using the same statistical protocol described in Section~\ref{sec:exp_setup}.

\subsection{Ablation and Component Contribution}

A central question is how much each component of SafeMind contributes to overall safety and performance. Following standard ablation practice in learning-based control and safe RL \cite{brunke2022safelearning,cheng2019rlcbf}, we remove major modules individually---the stochastic barrier correction term, the risk-aware tightening term governed by $\kappa$, the meta-adaptive update mechanism, and the semantic encoder---and evaluate changes in SVR, tracking accuracy, and energy. To further isolate the role of online adaptation from the role of risk sensitivity itself, we additionally evaluate a fixed-parameter variant in which $\alpha$ and $\kappa$ are held constant at their nominal values throughout each rollout.

Table~\ref{tab:ablation_full} reports results averaged over ten terrain categories and six semantic tasks. Removing either the stochastic correction or the risk-aware term leads to a dramatic increase in SVR (from $0.58\%$ to $3.74\%$ and $5.92\%$, respectively), with only modest changes in tracking error. This suggests that the stochastic modeling and risk-sensitive tightening primarily shape the safety envelope rather than the nominal tracking behavior.

\begin{table}[ht]
\centering
\scriptsize
\caption{Extended ablation across all simulation domains. Continuous valued metrics are reported as mean $\pm$ 95\% CI unless otherwise stated. Proportion-based metrics are reported as point estimates.}
\label{tab:ablation_full}
\begin{tabular}{lccc}
\toprule
\textbf{Configuration}
& \textbf{SVR (\%)}
& \textbf{RMSE (cm)}
& \textbf{Energy} \\
\midrule
Full SafeMind
& \textbf{0.58}
& \textbf{2.9 $\pm$ 0.2}
& \textbf{0.96 $\pm$ 0.03} \\
w/o Stochastic Term
& 3.74
& 3.3 $\pm$ 0.2
& 1.04 $\pm$ 0.04 \\
w/o Risk Tightening $(\kappa=0)$
& 5.92
& 3.2 $\pm$ 0.2
& 1.03 $\pm$ 0.04 \\
w/o Meta-Adaptation
& 1.48
& 3.1 $\pm$ 0.2
& 1.07 $\pm$ 0.04 \\
w/o Semantic Encoder
& 2.61
& 3.5 $\pm$ 0.3
& 1.01 $\pm$ 0.03 \\
Fixed $(\alpha,\kappa)$
& 1.83
& 3.1 $\pm$ 0.2
& 1.02 $\pm$ 0.03 \\
\bottomrule
\end{tabular}
\end{table}

Beyond aggregate statistics, the ablation reveals structural differences. Without the stochastic correction, the controller underestimates the degradation of barrier values under uncertain contacts, effectively creating optimistic ``false margins'' similar to deterministic CBF failures reported in stochastic settings.\cite{fridovich2020stochastic,clark2021cbf} Disabling $\kappa$ causes SafeMind to maintain nominal-risk envelopes even when uncertainty grows, leading to delayed reactions during terrain changes. Meta-adaptation primarily affects responsiveness: it reduces recovery time after perturbations and terrain transitions by more than one-third, but has a smaller effect on steady-state SVR. Finally, removing the semantic encoder increases violations in tasks where constraints depend on high-level scene structure, indicating that SafeMind benefits from a hierarchical integration of semantic reasoning and low-level safety control.\cite{brohan2023rt1,shah2023lmnav} The fixed-parameter variant further shows that online adaptation of $\alpha$ and $\kappa$ provides measurable benefit beyond simply choosing a conservative static setting, especially during rapid terrain transitions and uncertainty surges.

\subsection{Sensitivity and Parameter-Scaling Analysis}

To study robustness under model variation, we systematically scale key physical and sensing parameters: mass and inertia, friction, perception latency, and control frequency. Sensitivity analysis is widely used in robust MPC and stochastic control to validate controller resilience to modeling errors.\cite{mesbah2016stochasticmpc,rawlings2009mpc} Table~\ref{tab:sensitivity} reports SVR, RMSE, and energy when each parameter is perturbed independently around its nominal value.

\begin{table}[ht]
\centering
\scriptsize
\caption{Sensitivity of SafeMind to parameter scaling. Continuous valued metrics are reported as mean $\pm$ 95\% CI unless otherwise stated. Proportion-based metrics are reported as point estimates.}
\label{tab:sensitivity}
\begin{tabular}{lccc}
\toprule
\textbf{Parameter Scaling} & \textbf{SVR (\%)} & \textbf{RMSE (cm)} & \textbf{Energy} \\
\midrule
Mass $+15\%$               & 0.72 & 3.1 $\pm$ 0.2 & 0.99 $\pm$ 0.03 \\
Mass $-15\%$               & 0.69 & 3.0 $\pm$ 0.2 & 0.98 $\pm$ 0.03 \\
Friction $-40\%$           & 1.31 & 3.4 $\pm$ 0.2 & 1.02 $\pm$ 0.03 \\
Latency $+20$\,ms          & 1.12 & 3.7 $\pm$ 0.3 & 1.05 $\pm$ 0.04 \\
Control Frequency 100\,Hz  & 1.84 & 3.3 $\pm$ 0.2 & 1.11 $\pm$ 0.04 \\
Control Frequency 50\,Hz   & 3.02 & 3.8 $\pm$ 0.3 & 1.20 $\pm$ 0.05 \\
\bottomrule
\end{tabular}
\end{table}

Two observations emerge. First, SafeMind remains robust under substantial mass and inertia perturbations, with SVR staying below $1.5\%$ even when friction is reduced by $40\%$. This indicates that risk-aware tightening compensates for dynamics mismatch: as friction drops or latency increases, the meta-adaptive update raises $\kappa$ and narrows the feasible safety tube.\cite{howell2022cbf} Second, although SVR increases at lower control frequencies, SafeMind maintains comparatively low violation rates and moderate energy growth, suggesting that its barrier shaping provides intrinsic stability even under degraded actuation bandwidth. Deterministic baselines, by contrast, exhibit superlinear growth in violation rates under similar low-frequency stress (cf. Section~\ref{sec:exp_setup}). These trends also clarify the scope of the method: SafeMind remains effective under sizable but bounded perturbations, yet the gradual increase in SVR at very low control rates indicates that extreme discretization and bandwidth limits can eventually violate the assumptions underlying the mode-wise stochastic safety model.

\subsection{Risk Sensitivity Dynamics}

A distinguishing feature of SafeMind is the dynamic adaptation of the risk parameter $\kappa(\xi)$ in response to evolving uncertainty and environmental cues. Similar risk-sensitive adaptation has been explored in distributional and CVaR-based RL, but typically at the value-function level rather than at the level of state-wise safety constraints.\cite{chow2015cvar,dabney2018qr} Here, we directly inspect the temporal evolution of $\kappa$ during terrain transitions.

Table~\ref{tab:kappa_stats} summarizes statistics of $\kappa$ across several prototypical terrain changes. We observe a consistent pattern: upon entering uncertain terrain such as slippery or rubble regions, $\kappa$ exhibits a sharp ``risk surge'' within a few control cycles, followed by a slower ``risk relaxation'' as contact stabilizes.

\begin{table}[ht]
\centering
\caption{Statistics of risk sensitivity $\kappa$ across terrain transitions. Mean and Peak are reported as mean $\pm$ 95\% CI; Std denotes the empirical standard deviation of $\kappa$ over the transition window.}
\label{tab:kappa_stats}
\begin{tabular}{lccc}
\toprule
\textbf{Terrain Transition} & \textbf{Mean $\kappa$} & \textbf{Std} & \textbf{Peak} \\
\midrule
Flat $\rightarrow$ Slippery & 2.48 $\pm$ 0.14 & 0.91 & 4.83 $\pm$ 0.27 \\
Soft $\rightarrow$ Rubble   & 3.07 $\pm$ 0.18 & 1.12 & 5.14 $\pm$ 0.31 \\
Rubble $\rightarrow$ Gap    & 2.81 $\pm$ 0.16 & 0.97 & 4.55 $\pm$ 0.28 \\
Gap $\rightarrow$ Flat      & 1.21 $\pm$ 0.09 & 0.41 & 2.19 $\pm$ 0.15 \\
\bottomrule
\end{tabular}
\end{table}

The asymmetric time profile---fast rise and slow decay---implements a conservative hysteresis: SafeMind reacts quickly to rising uncertainty, but relaxes risk sensitivity more cautiously. This hysteresis reduces the likelihood of over-optimistic barrier relaxation after short-lived disturbances, a failure pattern often reported in deterministic safety filters.\cite{fridovich2020stochastic} Importantly, the bounded and smoothed updates prevent unstable parameter oscillations, supporting the practical claim that online adaptation improves resilience without destabilizing the inner-loop controller.

\subsection{Constraint Activation and Safety Geometry}

To understand how SafeMind distributes safety effort across multiple constraints, we measure the activation frequency of individual barrier functions $h_i$ across terrains and tasks. The resulting statistics, shown in Table~\ref{tab:activation}, reveal how the ``safety geometry'' changes with environment.

\begin{table}[ht]
\centering
\caption{Constraint activation frequencies (percentage of active cycles). All values are reported as point estimates across repeated runs.}
\label{tab:activation}
\begin{tabular}{lcccc}
\toprule
\textbf{Constraint} & T2 & T4 & T5 & T8 \\
\midrule
Foot Clearance      & 22.3 & 18.1 & 13.8 & 31.4 \\
Body Tilt           & 11.8 & 33.2 & 24.9 & 29.5 \\
Joint Limits        & 4.3  & 5.9  & 4.8  & 6.4 \\
Semantic Boundaries & 0.0  & 2.7  & 4.6  & 7.1 \\
\bottomrule
\end{tabular}
\end{table}

On uneven and rubble terrains, foot-clearance constraints dominate, whereas on soft and slippery terrains, body-tilt constraints are more frequently active, reflecting the need to regulate pitch and roll under compliance and low friction.\cite{hwangbo2019learning,miki2022learning} The nonzero activation of semantic constraints on rubble (T8) indicates that SafeMind can maintain semantic region constraints even in physically challenging conditions, illustrating a tight coupling between high-level intent and low-level safety enforcement. At the same time, the comparatively lower activation rate of semantic boundaries relative to core physical constraints is consistent with the intended arbitration policy: semantic cues influence safety geometry without overriding contact, balance, or actuation feasibility when conflicts arise.

\subsection{Robustness Stress Testing}

We further evaluate SafeMind under rare-event and adversarial disturbances: large external pushes (50\,N), sudden friction collapse (down to $\mu<0.1$), temporary perception dropout (300\,ms), and abrupt 10\,cm height steps. Similar stress tests are common in robustness studies of quadruped locomotion and safety filters.\cite{lee2020learning,howell2022cbf} Table~\ref{tab:stress_tests} reports aggregated SVR, episode-level failure rate, and recovery time.

\begin{table}[ht]
\centering
\scriptsize
\caption{Aggregated results under rare-event stress tests. Recovery time is reported as mean $\pm$ 95\% CI. Proportion-based metrics are reported as point estimates. Failure rate is reported as observed episode-level failure percentage with count in parentheses.}
\label{tab:stress_tests}
\begin{tabular}{lccc}
\toprule
\textbf{Stress Scenario} & \textbf{SVR (\%)} & \textbf{Failure Rate} & \textbf{Recovery Time (s)} \\
\midrule
50\,N Push              & 1.14 & 0.0 (0/200) & 1.25 $\pm$ 0.09 \\
Friction Collapse       & 2.38 & 0.0 (0/200) & 1.91 $\pm$ 0.13 \\
300\,ms Perception Drop & 1.07 & 0.0 (0/200) & 1.48 $\pm$ 0.10 \\
10\,cm Drop-off         & 1.44 & 0.0 (0/200) & 1.72 $\pm$ 0.12 \\
\bottomrule
\end{tabular}
\end{table}

In all evaluated scenarios, SafeMind avoids catastrophic failures, whereas Safety--MPC and deterministic CBF baselines exhibit non-negligible episode-level failure rates (3--8\%) under the same conditions. The results indicate that stochastic barrier modeling and meta-adaptive risk shaping allow SafeMind to ``absorb'' unexpected perturbations without leaving the safe set. Nevertheless, the friction-collapse case remains the most challenging, suggesting that severe contact-model mismatch is still a dominant source of residual risk even for uncertainty-aware safety filters.

\subsection{Long-Horizon Stability and Drift}

Preventing error accumulation during long-term operation is crucial for field deployment.\cite{rudin2022learning,brunke2022safelearning} We therefore evaluate SafeMind on 600\,s episodes with periodic disturbances and monitor the cumulative probability of safety violations over time. Table~\ref{tab:long_horizon} compares accumulated violation probabilities for SafeMind and several baselines.

\begin{table}[ht]
\centering
\caption{Long-horizon accumulated violation probability. All values are reported as mean $\pm$ 95\% CI unless otherwise stated.}
\label{tab:long_horizon}
\begin{tabular}{lcccc}
\toprule
\textbf{Controller} & 150\,s & 300\,s & 450\,s & 600\,s \\
\midrule
Nominal MPC        & 4.2\% & 7.3\% & 9.7\% & 12.4\% \\
Safety--MPC        & 1.9\% & 3.1\% & 4.8\% & 5.6\% \\
BarrierNet         & 1.6\% & 2.8\% & 3.7\% & 4.2\% \\
\textbf{SafeMind}  & \textbf{0.41\%} & \textbf{0.58\%} & \textbf{0.65\%} & \textbf{0.71\%} \\
\bottomrule
\end{tabular}
\end{table}

Deterministic baselines exhibit monotonic growth of accumulated violations, consistent with a gradual optimistic bias in barrier evolution. In contrast, SafeMind stabilizes quickly and maintains a nearly flat violation curve, indicating that the stochastic correction term counteracts bias accumulation over long horizons. This behavior is particularly important in realistic deployments where robots operate for many minutes or hours without human supervision. The near-saturation of the violation curve also suggests that the bounded adaptation mechanism does not accumulate destabilizing drift over long operation windows in the evaluated regimes.

\subsection{Safety--Performance Pareto Frontier}

Finally, we analyze the intrinsic trade-off between safety and agility by sweeping the nominal risk parameter $\kappa$ while keeping other components fixed. Similar safety--performance trade-off analyses appear in constrained and risk-sensitive RL,\cite{chow2015cvar,achiam2017cpo} but here we examine the trade-off at the level of barrier-enforced state trajectories.

Table~\ref{tab:pareto} reports SVR, energy, and RMSE for several fixed values of $\kappa$. As $\kappa$ increases, SafeMind becomes more conservative, reducing SVR at the cost of slightly higher energy and tracking error.

\begin{table}[ht]
\centering
\caption{Safety--performance Pareto frontier for varying $\kappa$. Continuous valued metrics are reported as mean $\pm$ 95\% CI across repeated trials. Proportion-based metrics are reported as point estimates.}
\label{tab:pareto}
\begin{tabular}{cccc}
\toprule
$\kappa$ & \textbf{SVR (\%)} & \textbf{Energy} & \textbf{RMSE (cm)} \\
\midrule
0.5  & 1.82 & 0.91 $\pm$ 0.03 & 2.4 $\pm$ 0.2 \\
1.5  & 0.93 & 0.95 $\pm$ 0.03 & 2.7 $\pm$ 0.2 \\
3.0  & 0.58 & 0.96 $\pm$ 0.03 & 2.9 $\pm$ 0.2 \\
5.0  & 0.41 & 1.04 $\pm$ 0.04 & 3.3 $\pm$ 0.2 \\
\bottomrule
\end{tabular}
\end{table}

The resulting curve exhibits a clear ``knee'': moderate $\kappa$ values (around 3.0) offer substantial safety improvements with only marginal energy overhead, whereas further increases lead to diminishing returns. SafeMind's meta-adaptive mechanism operates near this knee in most scenarios, adjusting $\kappa$ upward only when uncertainty grows, and thereby remaining close to the empirical safety--performance Pareto frontier.

\subsection{Failure Case Analysis}

Although SafeMind shows strong overall performance, minor failure modes persist. We group the observed failures into four categories. First, on highly compliant or abruptly changing terrain, mild pitch oscillations can appear when contact uncertainty is underestimated over several consecutive steps. Second, on extreme rubble or partial footholds, occasional foot scuffing occurs when local barrier curvature becomes difficult to estimate reliably near sharp geometric discontinuities. Third, under temporary semantic ambiguity or low-confidence perception, SafeMind may become briefly over-conservative, slowing down rather than violating hard physical constraints. Fourth, in rare cases involving simultaneous activation of multiple hard constraints, the controller approaches the edge of QP feasibility and exhibits short near-boundary excursions.

These issues arise when compounded uncertainties lead to regions where local barrier curvature becomes difficult to estimate. Importantly, such violations remain subcritical, appearing as brief near-boundary excursions rather than catastrophic failures, consistent with graceful degradation in safe control literature \cite{brunke2022safelearning}. These failure cases are consistent with the mode-wise scope of the proposed theory: they occur primarily when contact transitions, perception degradation, or severe environment shifts stress the assumptions used in the stochastic safety model. They therefore highlight practical limitations of the current implementation rather than contradictions of the nominal safety construction.

Collectively, ablation, sensitivity, and long-horizon evaluations indicate that SafeMind’s robustness stems from the coordinated interaction of uncertainty modeling, adaptive risk shaping, and differentiable multi-constraint optimization, rather than rigid worst-case conservatism.

\section{Discussion and Limitations}

Although SafeMind demonstrates strong safety, adaptability, and efficiency across both simulation and hardware experiments, several limitations should be noted. First, the theoretical guarantees are derived under a mode-wise stochastic formulation and therefore apply most directly within continuous contact phases. In hybrid quadruped locomotion, contact switches and impact events are handled in practice through conservative margins, fast replanning, and low-level stabilization rather than through a complete global hybrid-system proof.

Second, the effectiveness of the proposed safety layer depends on the quality of uncertainty estimation and semantic perception. If covariance is substantially underestimated, or if semantic context is severely misclassified, the resulting barrier modulation may become insufficiently conservative or temporarily over-conservative. While the experiments show graceful degradation in such cases, these effects remain an important practical limitation.

Third, although SafeMind generalizes across the two quadruped platforms studied here and remains robust under morphology perturbations such as added payload and partial leg impairment, extension to substantially different robot morphologies or higher-dimensional systems is not automatic. Such transfer requires task-appropriate barrier construction, contact-aware uncertainty modeling, and solver configurations consistent with the new dynamics.

Fourth, any SystemC- or virtual-prototype-based backend discussed as part of the broader development workflow remains conceptual in the current version. It is not a fully validated component of the deployed SafeMind control stack, and the experimental results reported in this paper do not rely on end-to-end verification through that backend.

Finally, the current semantic integration mechanism is designed so that physical safety constraints retain priority over semantic preferences. This avoids unsafe behavior under conflicting instructions, but may also lead to conservative task execution in environments with ambiguous or low-confidence semantic cues. A more principled confidence-aware arbitration mechanism remains an important direction for future work.
\section{Conclusion}

We presented \textbf{SafeMind}, a stochastic, semantic-aware, and differentiable safety-control framework that integrates variance-aware barrier modeling, risk-sensitive optimization, and meta-adaptive parameter shaping. Extensive simulation and hardware experiments demonstrate substantial reductions in safety violation rates while maintaining or improving tracking accuracy and energy efficiency, particularly under abrupt terrain changes and severe uncertainty. Additional analyses show that SafeMind’s robustness arises from the synergy among stochastic correction, adaptive risk modulation, and semantic grounding. These results highlight probabilistic safety reasoning with differentiable control as a promising paradigm for reliable real-world legged locomotion.
\bibliographystyle{IEEEtran} 
\bibliography{bib}           
\newpage
\begin{IEEEbiography}[{\includegraphics[width=1in,height=1.25in,clip,keepaspectratio]{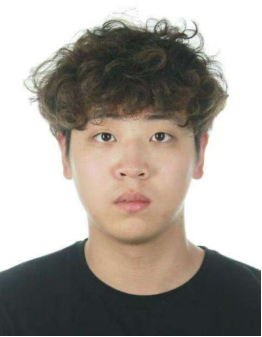}}]{Zukun Zhang} was born in Shanxi Province, China. He received the B.S. degree in Mechanical Design, Manufacturing and Automation from Taiyuan University of Technology. He is currently pursuing his studies at The University of Hong Kong, Pokfulam, Hong Kong.
His interests include reading and sports. He can be contacted at zhangzukun2025@126.com.
\end{IEEEbiography}

\begin{IEEEbiography}[{\includegraphics[width=1in,height=1.25in,clip,keepaspectratio]{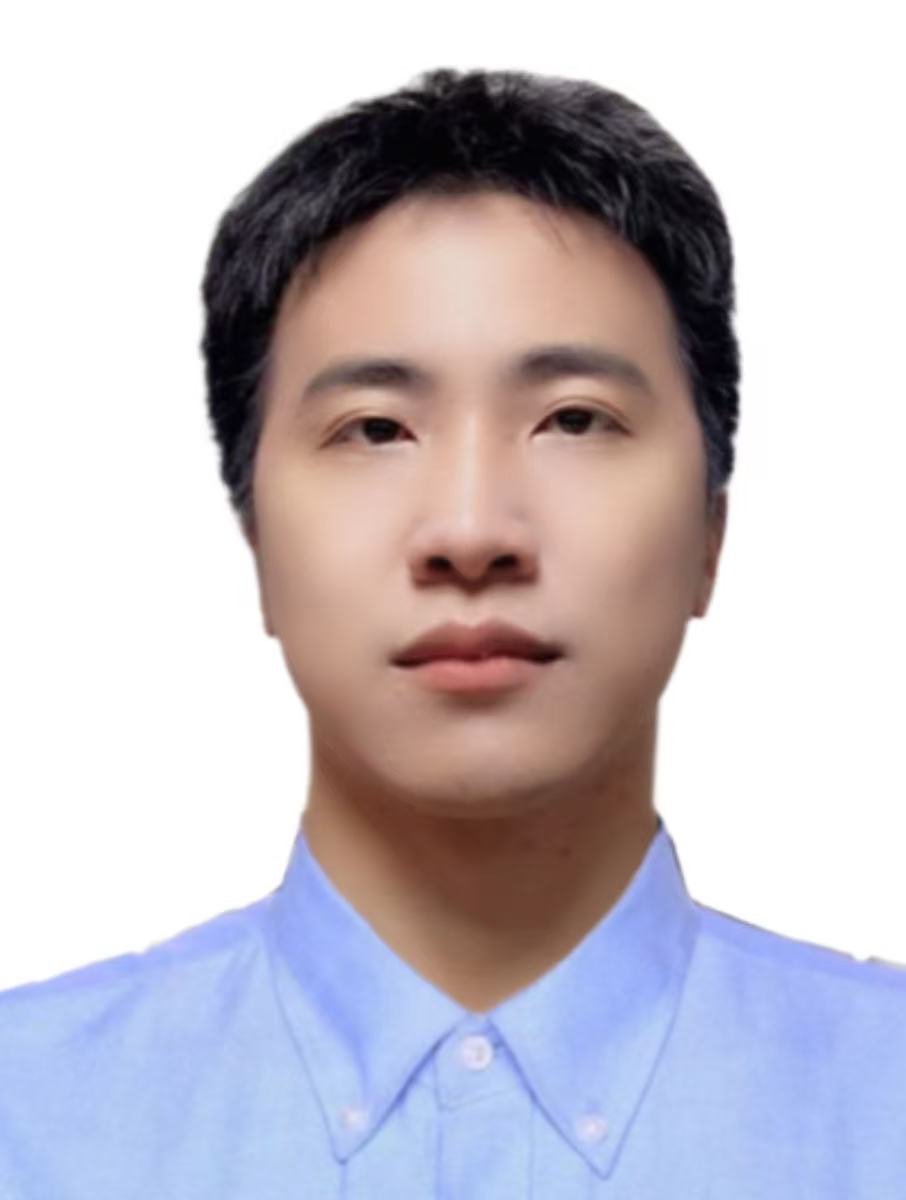}}]{Kai Shu} was born in Zhejiang Province, China. He earned his bachelor’s degree in Electronic Information Engineering from Xidian University, followed by a master’s degree in Computer Science from the University of Southern California. After graduation, he joined Alibaba Group, where he has been engaged in algorithm research and software development. His professional and research interests focus on Computer Vision (CV), Natural Language Processing (NLP), and AI for Science (AI4Science). He is particularly interested in applying advanced artificial intelligence techniques to solve complex real-world and scientific problems.His interests include reading and sports. He can be contacted at kaishu.cs@gmail.com.
\end{IEEEbiography}

\begin{IEEEbiography}[{\includegraphics[width=1in,height=1.25in,clip,keepaspectratio]{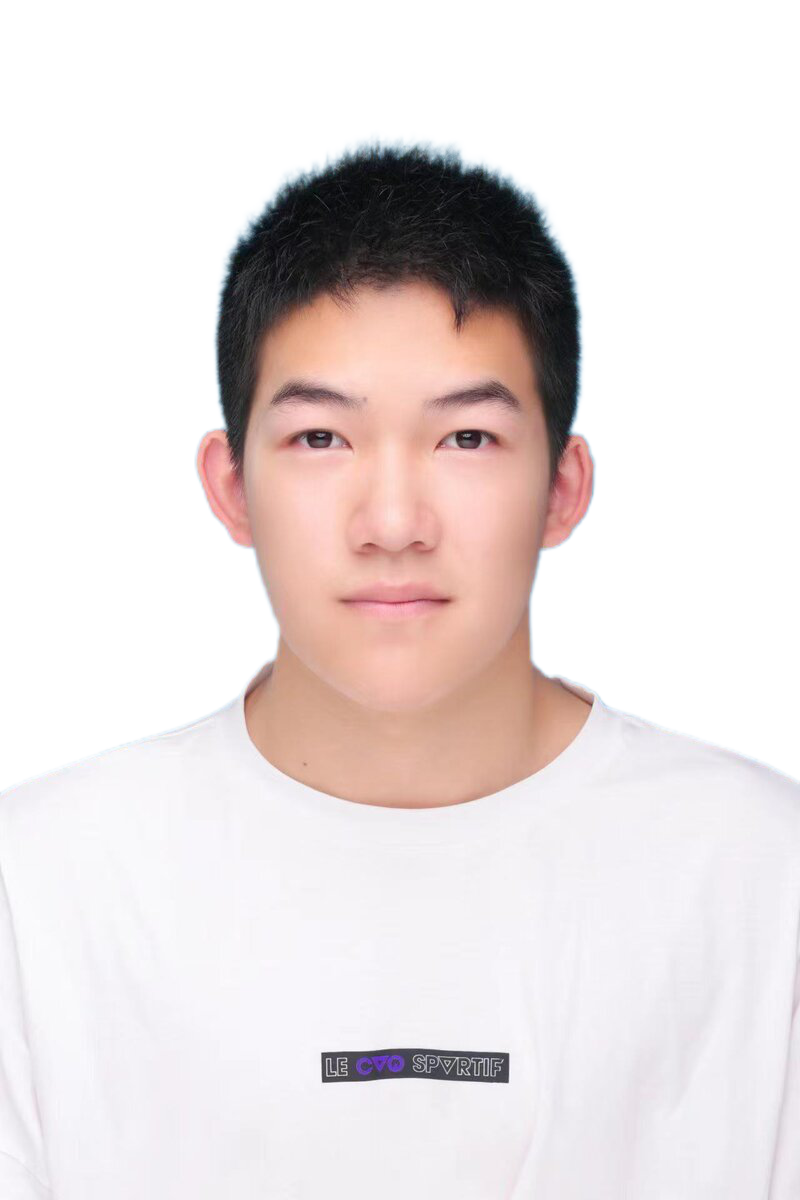}}]{Mingqiao Mo} was born in China. He is currently an intern at the University of the Chinese Academy of Sciences. He joined in January 2024. He has previously held research internship positions at Tsinghua University, the University of California, Los Angeles, and Cornell University. His research experience spans multiple leading academic institutions from 2023 to 2025.His research interests include large language model (LLM) reasoning, diffusion models, and computer vision. He has authored and co-authored several research papers in the areas of compiler optimization, robust representation learning, and multimodal reasoning, with submissions to top-tier conferences such as ACL, AAMAS, ICLR, and WACV. His recent work focuses on leveraging large language models for program optimization and robust code representation learning. 
\end{IEEEbiography}
\EOD
\end{document}